\documentclass[10pt,twocolumn,letterpaper]{article}
\usepackage[accsupp]{axessibility}  

\usepackage{iccv}
\usepackage{times}
\usepackage{epsfig}
\usepackage{graphicx}
\usepackage{amsmath}
\usepackage{amssymb}

\usepackage[format=plain,labelformat=simple,labelsep=period,font=small,compatibility=false]{caption}
\usepackage{caption}
\usepackage{bbm}
\usepackage{booktabs}
\usepackage[hang,flushmargin]{footmisc}

\usepackage{tablefootnote}
\usepackage{multirow}
\usepackage{wrapfig}
\usepackage{colortbl}
\usepackage{enumitem}
\usepackage{graphicx}
\usepackage{amssymb}
\usepackage{pifont}
\usepackage{subcaption}
\usepackage{amsmath}
\usepackage{makecell}
\usepackage{xcolor}
\usepackage{listings}

\setlist[itemize]{align=parleft,left=0pt}

\definecolor{table-gray}{gray}{0.85}
\definecolor{light-gray}{gray}{0.8}
\definecolor{azure(colorwheel)}{rgb}{0.0, 0.5, 1.0}
\definecolor{nicegreen}{rgb}{0.0, 0.7, 0.1}
\definecolor{CuGray}{gray}{0.9}
\definecolor{pink}{cmyk}{0, 0.7808, 0.4429, 0.1412}
\definecolor{amethyst}{rgb}{0.6, 0.4, 0.8}
\definecolor{black}{rgb}{0.0, 0.0, 0.0}
\definecolor{emphasis}{rgb}{0.0, 0.5, 0.2}
\definecolor{deemphasis}{rgb}{0.85, 0.0, 0.1}
\definecolor{clova}{rgb}{0.24, 0.63, 0.33}
\definecolor{orange}{rgb}{1.0, 0.5, 0.0}

\renewcommand{\paragraph}[1]{\noindent\textbf{#1}\,\,}

\usepackage{xspace}

\def \alambicw {\includegraphics[width=0.015\linewidth]{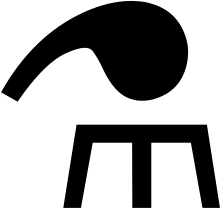}\xspace}

\def\distill{\alambicw}

\def\msa{\mathtt{MSA}}
\def\mlp{\mathtt{MLP}}

\def\norm{\mathtt{LN}}
\def\softmax{\mathtt{softmax}}

\def\Front{\mathtt{Front}}
\def\Mid{\mathtt{Mid}}
\def\End{\mathtt{End}}

\def\gs{\mathtt{CB}}
\def\gss{\mathtt{CB_S}}

\def\gsmul{\mathtt{CB}_{\texttt{gate}}}
\def\gshyb{\mathtt{CB}_{\texttt{hybrid}}}

\def\onedot{.\@\xspace}
\def\eg{\emph{e.g}\onedot} 
\def\ie{\emph{i.e}\onedot}

\def\etal{\emph{et al}\onedot}

\newcommand{\Sref}[1]{Sec.~\ref{#1}}

\newcommand{\Fref}[1]{Fig.~\ref{#1}}
\newcommand{\Tref}[1]{Table~\ref{#1}}

\newcommand{\ba}{{\mathbf{a}}}

\newcommand{\bp}{{\mathbf{p}}}

\newcommand{\bs}{{\mathbf{s}}}

\newcommand{\bx}{{\mathbf{x}}}

\newcommand{\bA}{\mathbf{A}}

\newcommand{\bJ}{\mathbf{J}}

\newcommand{\bS}{\mathbf{S}}

\newcommand{\bX}{\mathbf{X}}

\newcommand{\bLambda}{\mbox{\boldmath $\Lambda$}}

\newcommand{\Real}{\mathbb R}

\newcommand{\be}{\begin{eqnarray}}
\newcommand{\ee}{\end{eqnarray}}
\newcommand{\bee}{\begin{eqnarray*}}
\newcommand{\eee}{\end{eqnarray*}}

\newcommand{\matrixb}{\left[ \begin{array}}
\newcommand{\matrixe}{\end{array} \right]}

\newcommand{\cmark}{\ding{51}}%
\newcommand{\xmark}{\textcolor{light-gray}{\ding{55}}}%

\definecolor{codegreen}{rgb}{0,0.6,0}
\definecolor{codegray}{rgb}{0.5,0.5,0.5}
\definecolor{codepurple}{rgb}{0.58,0,0.82}
\definecolor{backcolour}{rgb}{0.95,0.95,0.92}

\lstdefinestyle{mystyle}{
  backgroundcolor=\color{backcolour}, commentstyle=\color{codegreen},
  keywordstyle=\color{magenta},
  numberstyle=\tiny\color{codegray},
  stringstyle=\color{codepurple},
  basicstyle=\ttfamily\footnotesize,
  breakatwhitespace=false,         
  breaklines=true,                 
  captionpos=b,                    
  keepspaces=true,                 
  numbers=left,                    
  numbersep=5pt,                  
  showspaces=false,                
  showstringspaces=false,
  showtabs=false,                  
  tabsize=2
}

\usepackage[normalem]{ulem}
\usepackage{colortbl}

\definecolor{azure(colorwheel)}{rgb}{0.0, 0.5, 1.0}
\definecolor{nicegreen}{rgb}{0.0, 0.7, 0.1}
\definecolor{CuGray}{gray}{0.9}
\definecolor{pink}{cmyk}{0, 0.7808, 0.4429, 0.1412}
\definecolor{amethyst}{rgb}{0.6, 0.4, 0.8}
\definecolor{black}{rgb}{0.0, 0.0, 0.0}

\definecolor{steelblue}{rgb}{0.27, 0.51, 0.7}
\definecolor{brickred}{rgb}{0.8, 0.25, 0.33}

\definecolor{customgray}{rgb}{0.9, 0.9, 0.9}

\usepackage[pagebackref,breaklinks,colorlinks,citecolor=blue,linkcolor=blue,bookmarks=false]{hyperref}
\usepackage[capitalize]{cleveref}
\crefname{section}{Sec.}{Secs.}
\Crefname{section}{Section}{Sections}
\Crefname{table}{Table}{Tables}
\crefname{table}{Tab.}{Tabs.}
\usepackage{xcolor}
\usepackage{listings}

\definecolor{mygray}{rgb}{0.9,0.9,0.9}

\iccvfinalcopy 

\def\iccvPaperID{8751} 

\ificcvfinal\fi

\begin{document}

\title{Scratching Visual Transformer's Back with Uniform Attention}

\author{
\begin{tabular}{c c c c}
     & Nam Hyeon-Woo$^{1}$\thanks{This work was done during N.~Hyeon-Woo's intern at NAVER AI Lab. Tae-Hyun Oh is in Department of Electrical Engineering and Grad.~School of Artificial Intelligence, POSTECH, and  joint affiliated with Institute for Convergence Research and Education in Advanced Technology, Yonsei University, Korea.} & Kim Yu-Ji$^{1}$ &  \\
    Byeongho Heo$^{2}$ & Dongyoon Han$^{2}$ & Seong Joon Oh$^{3}$ & Tae-Hyun Oh${}^{1*}$ \vspace{2mm}\\
\end{tabular}\\ 
    ${}^{1}$POSTECH, $^{2}$NAVER AI Lab, $^{3}$T\"{u}bingen University \\
}

\maketitle
\ificcvfinal\thispagestyle{empty}\fi

\begin{abstract}
    \vspace{-1em}
The favorable performance of Vision Transformers (ViTs) is often attributed to the multi-head self-attention ($\msa$), which enables global interactions at each layer of a ViT model.
Previous works acknowledge the property of long-range dependency for the effectiveness in $\msa$. In this work, we study the role of $\msa$ in terms of the different axis, density.
Our preliminary analyses suggest that the spatial interactions of learned attention maps are close to dense interactions rather than sparse ones. This is a curious phenomenon because dense attention maps are harder for the model to learn due to $\softmax$. We interpret this opposite behavior against $\softmax$ as a strong preference for the ViT models to include dense interaction. We thus manually insert the dense uniform attention to each layer of the ViT models to supply the much-needed dense interactions. We call this method Context Broadcasting, $\gs$. Our study demonstrates the inclusion of $\gs$ takes the role of dense attention and thereby reduces the degree of density in the original attention maps by complying $\softmax$ in $\msa$. We also show that, with negligible costs of $\gs$ (1 line in your model code and no additional parameters), both the capacity and generalizability of the ViT models are increased.
\vspace{-1em}

\end{abstract}


\begin{figure}
    \centering
    \includegraphics[width=0.95\linewidth]{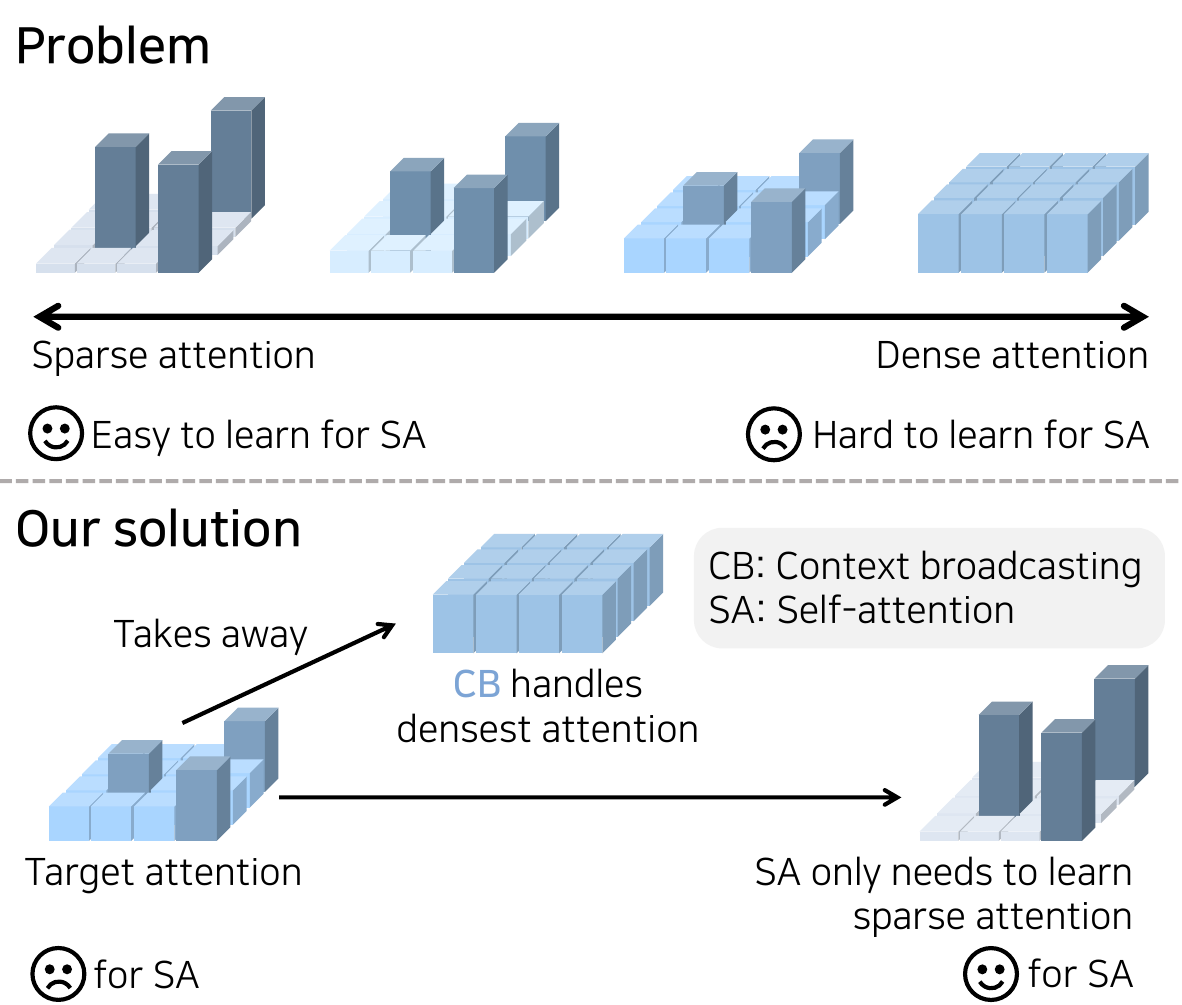}
    \vspace{-2mm}
    \caption{
    \textbf{Motivation of our work.}
    \textbf{Top:} dense attention is hard to learn with softmax, but self-attention tends to learn it more than sparse one.
    \textbf{Bottom:} we infuse dense attention
    explicitly, named $\gs$, to split the responsibility of interactions; the burden of interactions of self-attention is reduced. 
    Self-attention is now more likely to learn sparse interaction that is in favor of softmax.
    }
    \label{fig:teaser}
    \vspace{-1em}
\end{figure}

\section{Introduction}\label{sec:intro}

After the success of Transformers~\cite{transformer} in language domains, Dosovitskiy \etal~\cite{vit} have extended to Vision Transformers (ViTs) that operate almost identically to the Transformers but for computer vision tasks. Recent studies~\cite{vit,deit} have shown that ViTs achieve superior performance on image classification tasks. Further, the universal nature of ViTs' input has demonstrated its potential to multi-modal input extensions \cite{akbari2021vatt, girdhar2022omnivore, hu2021unit, vilt}.

The favorable performance is often attributed to the multi-head self-attention ($\msa$) in ViTs~\cite{vit, deit, wang2018non, carion2020end, strudel2021segmenter, raghu2021vision}, which facilitates long-range dependency\footnote{Long-range dependency is described in the literature with various terminologies: non-local, global, large receptive fields, etc.}. Specifically, $\msa$ is designed for long-range interactions of spatial information in all layers. This is a structurally contrasting feature with a large body of successful predecessors, convolutional neural networks (CNNs), which gradually increase the range of interactions by stacking many fixed and hard-coded local operations, \ie, convolutional layers. Raghu \etal~\cite{raghu2021vision} and Naseer \etal~\cite{NEURIPS2021_c404a5ad} have shown the effectiveness of the self-attention in ViTs for the global interactions of spatial information compared to CNNs.

Unlike previous works~\cite{raghu2021vision, NEURIPS2021_c404a5ad} that focused on the effectiveness of \textit{long-range dependency}, we study the role of \textit{density} in spatial attention. ``Long-range'' can be either ``sparse'' or ``dense''. We examine whether the learned attention of ViTs is dense or sparse. Our preliminary analysis based on the entropy measure suggests that the learned attention maps tend to be dense across all spatial locations. This is a curious phenomenon because denser attention maps are harder to learn by the softmax operation. Its gradients become larger (less stable) around denser attention maps. In other words, ViTs are trying hard to learn dense attention maps despite the difficulty of learning them through gradient descent.

While dense attention is unlikely to be learned via gradient descent, it is easy to implement it manually. We insert uniform attention explicitly, the densest form of attention, to confirm our observation of the effort of learning dense attention. We call our module Context Broadcasting ($\gs$). The module adds the averaged token to every individual token at intermediate layers. We find that when $\gs$ is added to ViT, $\gs$ reduces the degree of density in attention maps in all layers preserving the long-range dependency. $\gs$ takes over the role of the dense global aggregation from self-attention, as illustrated in \Fref{fig:teaser}. $\gs$ also makes the overall optimization for a ViT model easier and improves its generalization.

$\gs$ brings consistent gains in the image classification task on ImageNet~\cite{ILSVRC15,recht2019imagenet,beyer2020we} and the semantic segmentation task on ADE20K~\cite{adk20k, adk20k_2}. Overall, $\gs$ seems to help a ViT model divert its resources from learning dense attention maps to learning other informative signals. We also demonstrate that our module improves the Vision-Language Transformer, ViLT~\cite{vilt}, on a multi-modal 
task, VQAv2~\cite{balanced_vqa_v2}. Such benefits come with only negligible costs. Only 1 line of code needs to be inserted in your \texttt{model.py}. No additional parameters are introduced; only a negligible number of operations are. Our contributions are as follows:

\begin{itemize}
    \setlength\itemsep{-0.1em}
    \item Our observations of the dense interaction preference of ViTs but the learning difficulty from $\softmax$ 
    (\Sref{sec:motivation});
    \item A simple and effective modules, $\gs$ and $\gss$, for infusing dense interactions (\Sref{sec:ours});
    \item Phenomena for $\gs$ to divert 
    the capacity of $\msa$ for sparse interactions (\Sref{sec:abla});
\end{itemize}

\section{Related Work}\label{sec:related_work}

\paragraph{Transformers.}
Since the seminal work of the Transformers~\cite{transformer}, it has been the standard architecture in the natural language processing (NLP) domain. 
Dosovitskiy \etal~\cite{vit} have pioneered the use of Transformers in the visual domain with their Vision Transformers (ViTs).
The way of ViTs work is almost identical to the original Transformers, where ViTs tokenize non-overlapping patches of the input image and apply the Transformers architecture on top.
The Transformers with multi-head self-attention ($\msa$) are especially appealing in computer vision because their non-convolutional neural architectures do not have conventional hard-coded operations, such as convolution and pooling with fixed local kernel sizes.
Cordonnier \etal~\cite{Cordonnier2020On} and Ramachandran \etal~\cite{ramachandran2019stand} corroborate that the expressiveness of $\msa$ even includes convolution.

There have been attempts to understand the algorithmic behaviors of ViTs, including $\msa$, by contrasting them with CNNs~\cite{raghu2021vision, Cordonnier2020On, NEURIPS2021_c404a5ad, park2022how, tuli2021convolutional}.
Raghu \etal~\cite{raghu2021vision} empirically demonstrate early aggregation of global information and much larger effective receptive fields~\cite{luo2016understanding} over CNNs.
Naseer \etal~\cite{NEURIPS2021_c404a5ad} show highly robust behaviors of ViTs against diverse nuisances, including occlusions, distributional shifts, adversarial and natural perturbations.
Intriguingly, they attribute those advantageous properties to large and flexible receptive fields by $\msa$ in ViTs and interactions therein.
Similarly, there have been studies that attribute the effectiveness of $\msa$ to global interaction in many visual scene understanding tasks
~\cite{deit, carion2020end, strudel2021segmenter, raghu2021vision, lin2021end, arnab2021vivit, jiyeon2023mindvps}.
Distinctively, we study the role of the density of the attention.

\paragraph{Attention module.}
The global context is essential to capture a holistic understanding of a visual scene~\cite{torralba2003contextual, rabinovich2007objects, shotton2009textonboost, wang2018non, cao2022gcnet}, which aids visual recognition.
To capture the global context, models need to be designed to have sufficiently large receptive fields to interact and aggregate all the local information.
Prior arts have proposed to enhance the interaction range of CNNs by going deeper \cite{vgg, resnet, huang2017densely, tan2019efficientnet} or by expanding the receptive fields
\cite{YuK15, dai2017deformable, wang2018non, liu2015parsenet, huang2019ccnet}.
Hu \etal~\cite{hu2018squeeze} squeeze spatial dimensions by
pooling to capture the global context.
Cao \etal~\cite{cao2022gcnet} notice that the attention map of the non-local block is similar regardless of query position and propose a global context block.

Our study focuses on the ViT architecture, which comprises concise layers and can serve as 
versatile usage, such as
unified multi-modal Transformers~\cite{akbari2021vatt, girdhar2022omnivore,hu2021unit, vilt}.
The receptive field of $\msa$ in ViTs inherently covers the entire input space, which may facilitate the learning of global interactions and the modeling of global context more efficiently than CNNs~\cite{mao2021transformer}.
However, the current global context modeling in ViTs may not be straightforward.
Our research presents a few indications that while self-attention favors learning dense global interactions, it is challenging to achieve this due to $\softmax$.
To ascertain the benefits of dense global interaction, we explicitly inject it and observe an improvement of performance. 
Moreover, we observe the allocation of $\msa$ capacities to better interactions.
\section{Method}\label{sec:method}

We first motivate \emph{the need for 
dense interactions for the ViTs} in \Sref{sec:motivation}.
Then, we propose a simple, lightweight module and a technique 
to inject explicitly dense interactions into ViTs in \Sref{sec:ours}.
Finally, we demonstrate how uniform attention affects to the ViT model in \Sref{sec:abla}.

\begin{figure}
\vspace{-5mm}
  \small
  \centering
  \begin{subfigure}[b]{0.8\linewidth}
    \centering
    \includegraphics[width=1.0\linewidth]{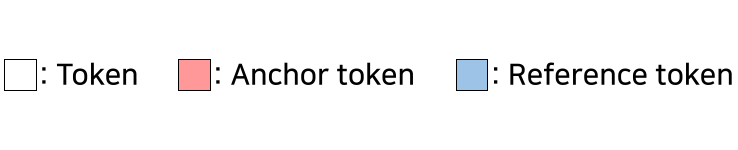}
    \vspace{-8mm}
  \end{subfigure}
  
   \begin{subfigure}[b]{0.30\linewidth}
    \centering
    \includegraphics[width=1\linewidth]{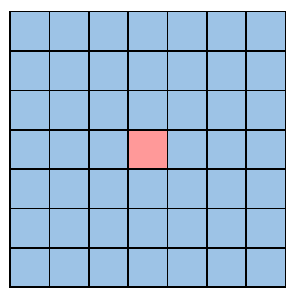}
    \caption{Dense global}
    \label{fig:DG}
 \end{subfigure}
 \hfill
 \begin{subfigure}[b]{0.30\linewidth}
    \centering
    \includegraphics[width=1\linewidth]{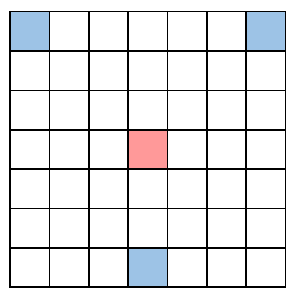}
    \caption{Sparse global}
    \label{fig:SG}
 \end{subfigure}
 \hfill
 \begin{subfigure}[b]{0.30\linewidth}
    \centering
    \includegraphics[width=1\linewidth]{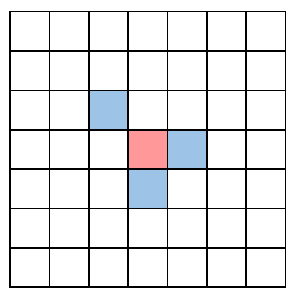}
    \caption{Sparse local}
    \label{fig:SL}
 \end{subfigure}
  \vspace{-2mm}
  \caption{\textbf{Type of spatial interactions.} We categorize the spatial interactions of self-attention into three types. 
  The anchor token interacts with reference tokens.}
  \label{fig:interaction}
  \vspace{-1em}
\end{figure}

\subsection{Motivation}\label{sec:motivation}
The self-attention operations let ViTs conduct spatial interactions without limiting the spatial range in every layer.
Long-range dependency, or global interactions, signifies connections that reach distant locations from the reference token. 
Density refers to the proportion of non-zero interactions across all tokens.
Observe that ``global'' does not necessarily mean ``dense'' or ``sparse'' because an attention map can be sparsely global.
We illustrate their difference in \Fref{fig:interaction}.
The question of interest is the type of interaction that self-attention learns.

Before delving into the study of density, we examine which multi-head self-attention ($\msa$) or multi-layer perceptron ($\mlp$) blocks further increase the capacity of the model.
Our preliminary observation highlights the benefit of studying $\msa$.
We then measure the layer-wise entropy of the attention to investigate the spatial interaction characteristics that ViTs prefer to learn.

\begin{figure}
    \centering
    \small
    \vspace{-4mm}
    \includegraphics[width=0.5\linewidth]{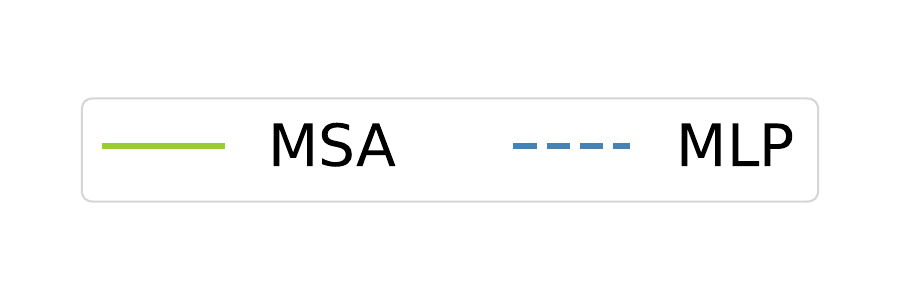}
    \vspace{-4mm}
    
    \hspace{-2mm}
    \begin{subfigure}[b]{0.5\linewidth}
         \centering
         \includegraphics[width=1.0\linewidth]{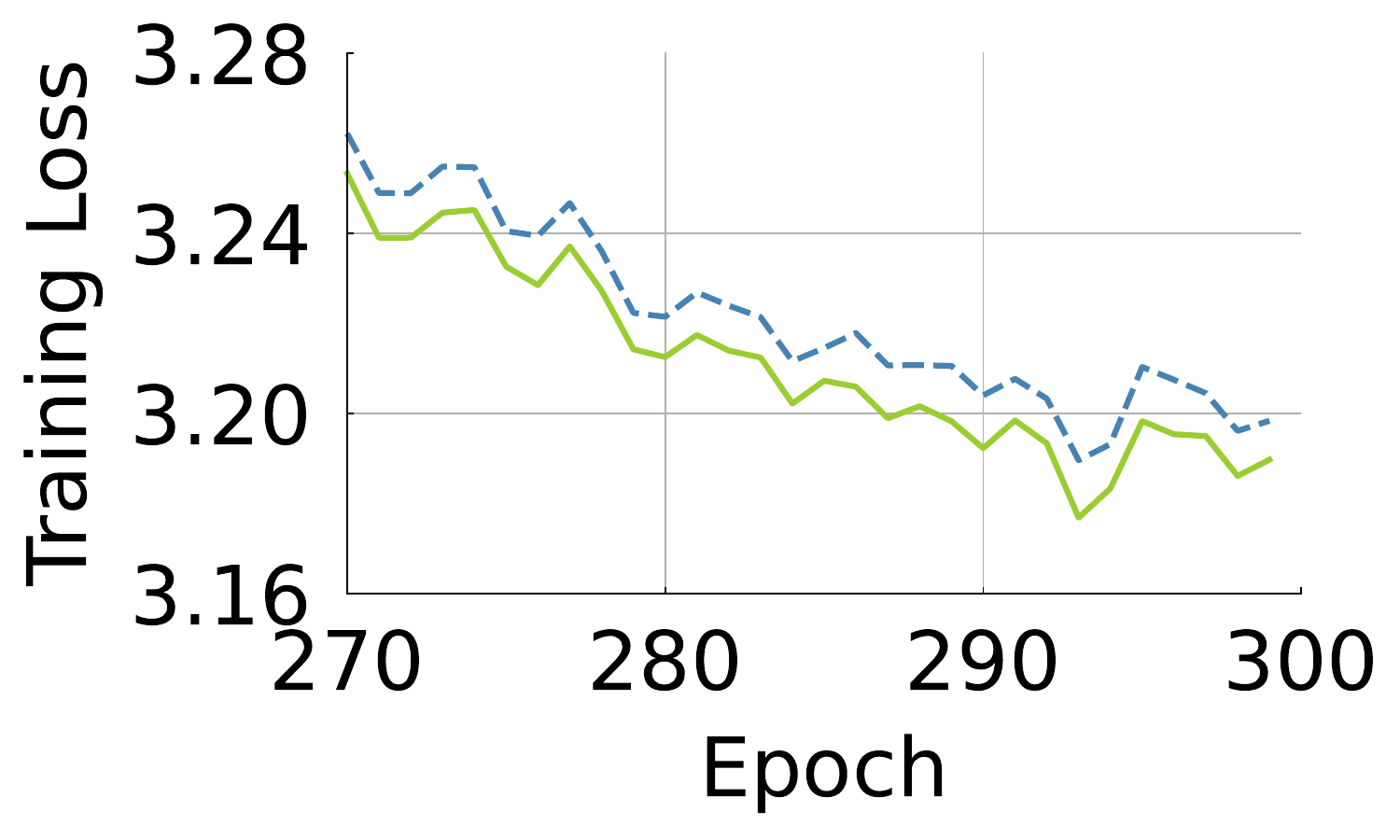}
         \vspace{-6mm}
         \caption{Training loss}
    \end{subfigure}
    \begin{subfigure}[b]{0.5\linewidth}
         \centering
         \includegraphics[width=1.0\linewidth]{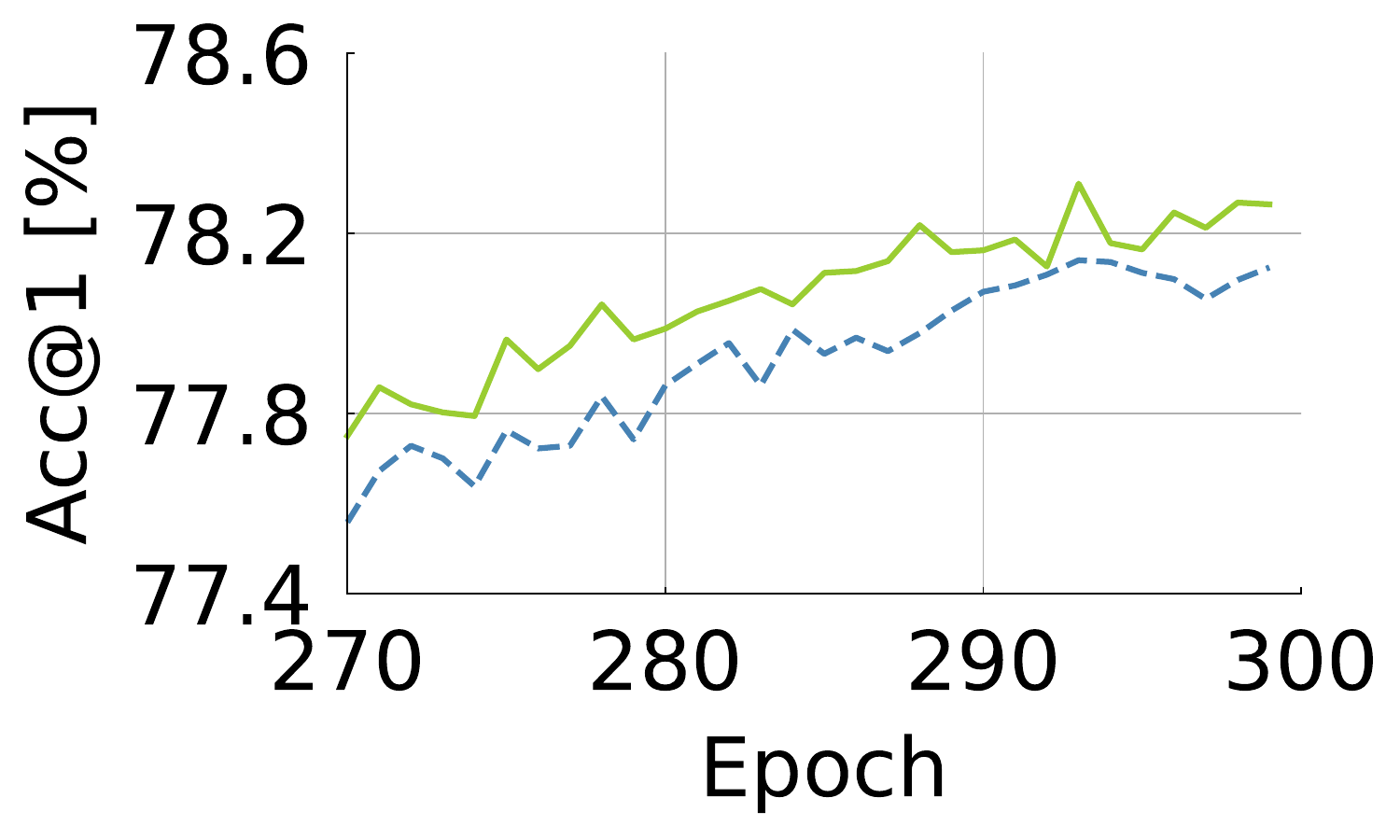}
         \vspace{-6mm}
         \caption{Top-1 accuracy}
    \end{subfigure}
    \vspace{-6mm}
    \caption{\small \textbf{Impact on the capacity of the ViT model with a single extra block.} Training loss and top-1 accuracy ($y$-axis) versus epochs ($x$-axis) of 8-depth ViT with additional $\msa$ and $\mlp$ blocks. The decrease in training loss and the increase in validation accuracy implies an increase in the model capacity.}
    \label{fig:toy_capacity}
    \vspace{-1em}
\end{figure}

\paragraph{MSA vs. MLP.}
$\msa$ and $\mlp$ in ViTs are responsible for spatial and channel interactions, respectively.
We examine adding which block, either $\msa$ or $\mlp$, increases the performance of ViTs more.
We train the eight-layer ViT on ImageNet-1K for 300 epochs with either an additional $\msa$ or $\mlp$ layer inserted at the last layer.
The additional number of parameters and FLOPs are nearly equal.
In \Fref{fig:toy_capacity}, we plot the training loss and top-1 accuracy.
We observe that the additional $\msa$ enables lower training loss and higher validation accuracy than the additional $\mlp$.
This suggests that, given a fixed budget in additional parameters and FLOPs, 
the ViT architecture seems to prefer to have extra spatial interactions rather than channel interactions.
It leads us to investigate the spatial interactions of $\msa$.

\paragraph{Which type of spatial interactions does $\msa$ learn?}
Here, we examine the types of spatial interactions that are particularly preferred by $\msa$.
Knowing the type of interactions will guide us on how we could improve attention performance.
While previous studies~\cite{raghu2021vision, NEURIPS2021_c404a5ad} have focused on the effectiveness of long-range dependency in $\msa$, we focus on the density in $\msa$.
We measure the dispersion of attention according to the depth through the lens of entropy.
Low entropy indicates that the attention is sparse, whereas high entropy suggests that the attention is dense.
Entropy provides a more objective view rather than relative and subjective measures such as visualization~\cite{ma2022close, he2021pruning}.

\begin{figure}
    \centering
    \small    
    \hspace{-2mm}
    \begin{subfigure}[b]{0.5\linewidth}
     \includegraphics[width=1\linewidth]{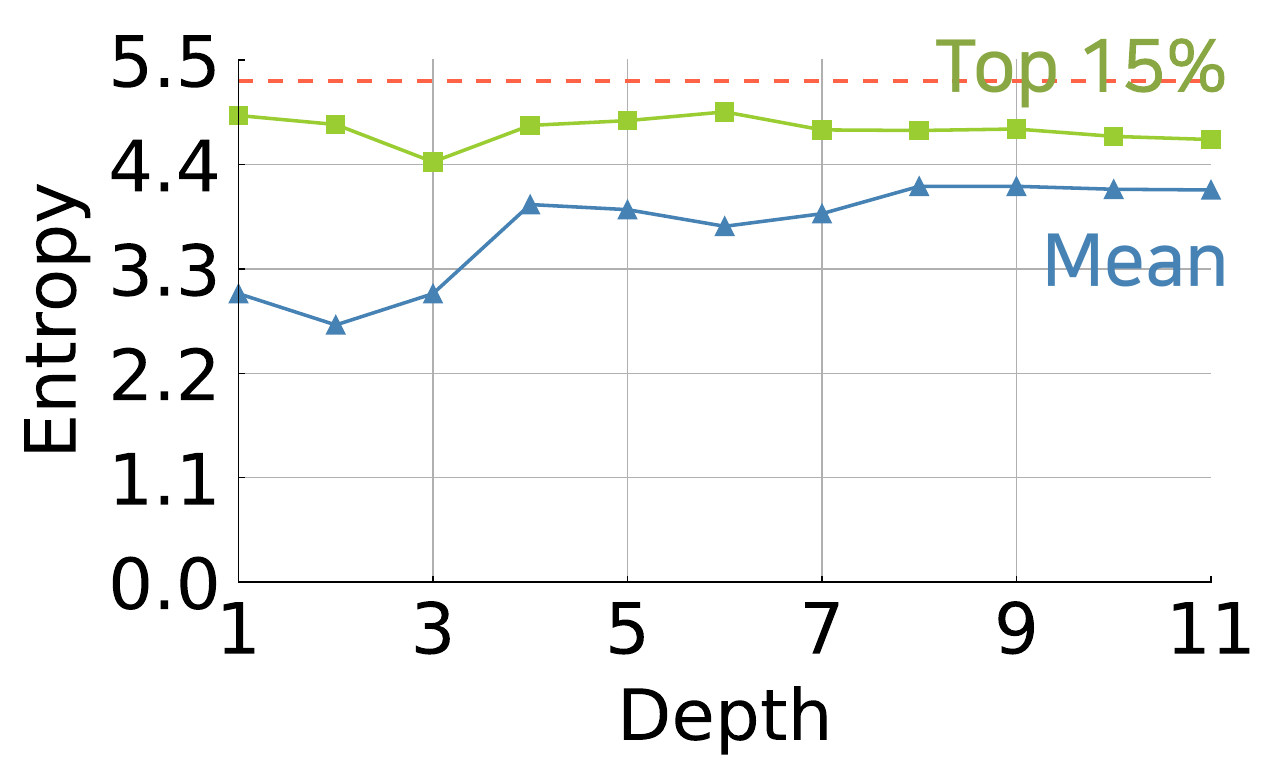}
     \vspace{-6mm}
     \caption{ViT-S}
     \label{fig:entro_topk_vit-s}
    \end{subfigure}
    \begin{subfigure}[b]{0.5\linewidth}
         \centering
         \includegraphics[width=1\linewidth]{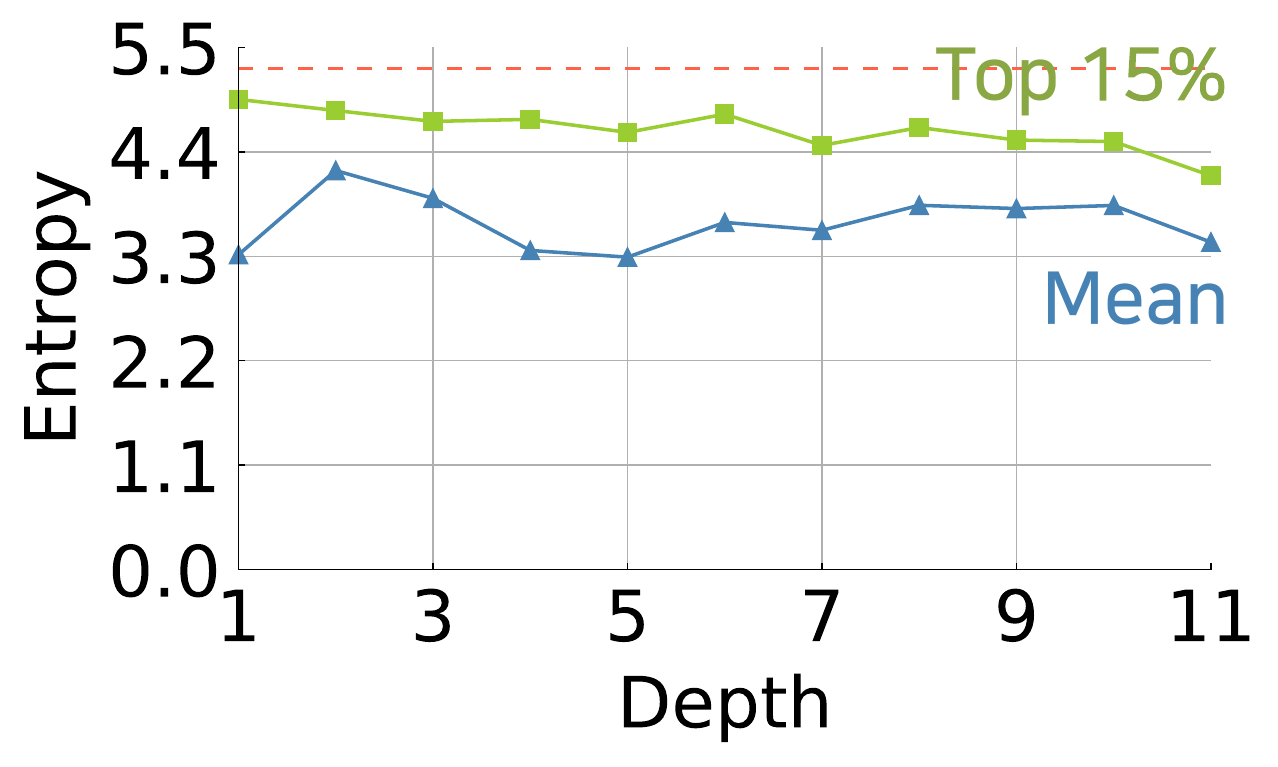}
         \vspace{-6mm}
         \caption{ViT-B}
         \label{fig:entro_topk_vit-b}
    \end{subfigure}
    \vspace{-6mm}
    \caption{\small \textbf{Entropy analysis.}
    We use pre-trained ViTs to measure layer-wise entropy.
    We plot the average and $15^{th}$ percentile of entropy values.
    The red dot line stands for the maximum entropy upper bound.
    }
    \label{fig:toy_entro_dist}
    \vspace{-1em}
\end{figure}

Figure \ref{fig:toy_entro_dist} shows the trends of the average and $15^{th}$ percentile entropy values across the heads and tokens for each $\msa$ layer in ViT-S/-B \cite{vit, deit}.
We observe that attention maps tend to have greater entropy values as high as 4.4 on average, towards the maximal entropy value, $-\sum \tfrac{1}{N}\log\tfrac{1}{N}  \approx 5.3$, where N is the number of tokens and 197.
The top 15\% of entropy values are much close to the maximal entropy value corresponding to uniform attention.
It is remarkable that a majority of the attention in ViTs has such high entropy values; it suggests that $\msa$ tends to learn the dense interactions.

\begin{figure}
    \centering
    \includegraphics[width=1.0\linewidth]{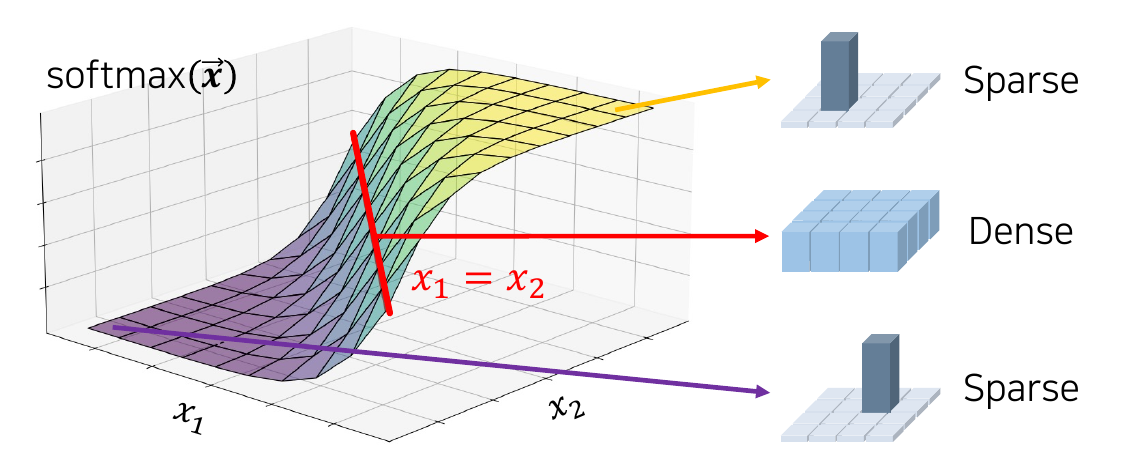}
    \vspace{-6mm}
    \caption{\textbf{Gradient around uniform attention.} 
    Softmax operation has high gradients around uniform attention ($x_1=x_2$).}
    \label{fig:softmax}
    \vspace{-1em}
\end{figure}

\paragraph{Steepest gradient around the uniform attention.}
The extreme form of dense interactions is the uniform distribution.
To examine the difficulty of finding the uniform distribution for the self-attention in $\msa$, 
we delve into the characteristics of the $\softmax$ function.
In a nutshell, we show that the gradient magnitude is the largest around the inputs inducing a uniform output. We further formalize this intuition below.
The self-attention
consists of the row-wise softmax operation $\bA=\sigma(\lambda\bS)\in\Real^{N\times N}$ where $\bS\in\Real^{N\times N}$ is the collection of dot products of queries and keys, possibly with a scale factor $\lambda>0$.
For simplicity, we consider the softmax over a single row: $\ba=\sigma(\lambda\bs)\in\Real^N$. The gradient of $\ba$ with respect to the input $\bs$ is $\bJ_{jk}:={\partial \ba_{j}}/{\partial \bs_{k}} = \lambda (\mathbbm{1}_{j=k}\ba_j -\ba_j\ba_k)$ for $1\leq j, k\leq N$.
We measure the magnitude of the gradient $\bJ\in\Real^{N\times N}$ using the nuclear norm $||\bJ||_*=\sum_{i=1}^N\nu_i$ where $\{\nu_i\}$ are the singular values of $\bJ$.
Note that $\bJ$ is a real, symmetric, and positive semi-definite matrix. Thus, the nuclear norm coincides with the sum of its eigenvalues, which in turn is the trace: $||\bJ||_* = \sum_j\lambda (\ba_j -\ba_j^2)$.
With respect to the constraint that $\sum_j \ba_j=1$ and $\ba_j\geq 0$ for all $j$, the nuclear norm $||\bJ||_*$ is maximal when $\ba_j=1/N$ for every $j$.
\Cref{fig:softmax} describes the $\softmax$ function in 2D input.
This shows that uniform attention with softmax can be easily broken by a single gradient step, meaning it is 
the most unstable type of attention to learn, 
in the optimization point of view.

\paragraph{Conclusion.}
We have examined the density of the interactions in the $\msa$ layers. 
We found that further spatial connections benefit ViT models more than further channel-wise interactions. 
$\msa$ layers tend to learn dense interactions with higher entropies. 
ViT's preference for dense interactions is striking, given the difficulty of learning dense interactions: the gradient for the $\msa$ layer is steeper with denser attention maps. 
This implies that dense attention maps are hard to learn but seem vital to ViTs.

\subsection{Explicitly Broadcasting the Context}\label{sec:ours}
We observe the curious phenomena: $\msa$ learns dense interaction, though it is unstable in terms of the gradient.
We decide to inject uniform attention because
(1) uniform attention is the densest attention and is unstable in terms of gradient view,
but (2) humans can supply uniform attention easily,
and (3) uniform attention requires no additional parameters and small computation costs.
We do this through the broadcasting context with the $\gs$ module.

\begin{figure*}
  \small
  \centering
  \begin{subfigure}[b]{0.4\linewidth}
    \centering
    \includegraphics[width=1\linewidth]{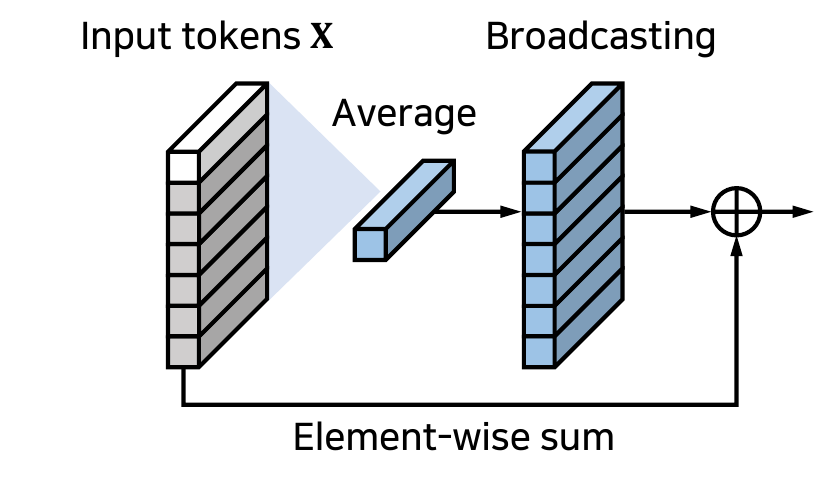}
    \vspace{-6mm}
    \caption{Our proposed $\gs$ module}
    \label{fig:our_module}
 \end{subfigure}
  \begin{subfigure}[b]{0.55\linewidth}
    \centering
    \vspace{2mm}
    \includegraphics[width=1.0\linewidth]{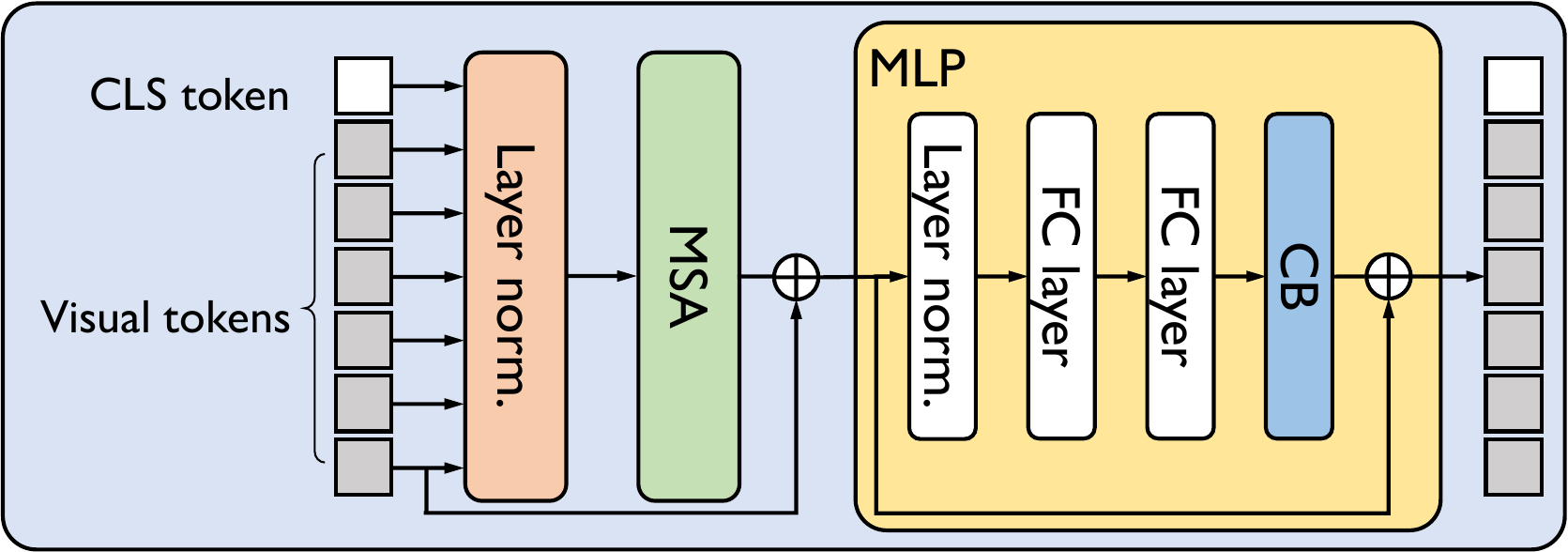}
    \vspace{-5mm}
    \caption{Our vision transformer}
    \label{fig:our_arch}
 \end{subfigure}
  \vspace{-1em}
  \caption{\small \textbf{Context Broadcasting ($\gs$) module.}
  (a) Our $\gs$ module broadcasts the context to each token.
  (b) The $\gs$ module is inserted at the end of the $\mlp$ block of the Vision Transformer (ViT) architectures. ViTs have other possible positions for our module, but we analyze that inserting at the end of $\mlp$ outperforms others.
  }
  \label{fig:ours}
  \vspace{-1em}
\end{figure*}

\paragraph{Context Broadcasting ($\gs$).}
Given a sequence of $N$ tokens $\bX \in \Real^{N \times d}$, our $\gs$ module supplies the averaged token back onto the tokens as follows:
\begin{equation}\label{eqn:ours}
    \gs(\bx_i) = \frac{\bx_i + \frac{1}{N}\sum_{j=1}^{N}\bx_j}{2}\quad\text{for every token $i$},
\end{equation}
where $\bx_i \in \Real^{d}$ is the $i^\text{th}$ token in $\bX$.
Figure~\ref{fig:our_module} illustrates our $\gs$ module.
The $\gs$ module is placed at the end of $\mlp$ block (See \cref{fig:our_arch}).
Our analysis in \Sref{subsec:where} shows that the insertion of $\gs$ increases the performance of ViTs regardless of its position. 
As we shall see, the performance increase is most significant when it is inserted after the $\mlp$ block.

\paragraph{Computational efficiency.}
The $\gs$ module is implemented with 1 line of code in deep learning frameworks like $\mathtt{PyTorch}$~\cite{NEURIPS2019_9015}, $\mathtt{Tensorflow}$~\cite{tensorflow}, and $\mathtt{JAX}$~\cite{jax}:

\noindent \colorbox{mygray}{\lstinline|X = 0.5 * X + 0.5 * X.mean(dim=1, keepdim=True)|}. It does not increase the number of parameters and incurs negligible additional operations for inference and training. 

\paragraph{$\gs$ with dimension scaling.}
Although we focus on the simplest form, we propose another variant of dense interaction $\gss$ by introducing a minimal number of parameters.
The proposed $\gs$ injects the dense interaction into all channel dimensions, but some channel dimensions of a token would require dense interaction, whereas others would not. We then introduce weights to scale the channels, $\bLambda \in \Real^{d}$, to infuse uniform attention selectively for each dimension as follows: $\gss(\bx_i) = \bx_i + \bLambda \odot \left( \frac{1}{N}\sum_{j=1}^{N}\bx_j \right)$ where $\odot$ is the element-wise product.
$\gss$ introduces few parameters: 0.02\% additional parameters for ViT-S.

\subsection{How Does Uniform Attention Affect ViT?}\label{sec:abla}
In the following experiments, we delve into the effect of uniform attention.
We train ViTs on ImageNet-1K during 300 epochs following the DeiT setting~\cite{deit}.

\paragraph{Does uniform attention help?}
To examine the effectiveness of the uniform attention, we inject the uniform attention in several ways to ViT-S as follows:
\textbf{(A)} We replace one of the multi-head self-attention heads to be $\gs$ which reduces the number of parameters corresponding to the replaced head,
\textbf{(B)} adjust the number of parameters of \textbf{(A)} to be comparable to the original ViT,
\textbf{(C)} append $\gs$ to $\msa$ as an extra head which increases the number of parameters,
and infuse $\gs$ and $\gss$.
\Tref{table:uniform_attention} shows the top-1 accuracy at ImageNet-1K.
In \textbf{(A)}, \textbf{(B)}, \textbf{(C)}, $\gs$, and $\gss$ improve the accuracy consistently.
The result explicitly tells us the broad benefits of injecting dense interactions into ViTs. 

\begin{table}
    \centering
    \small
    \tabcolsep=0.3cm
    \centering
    \begin{tabular}{c c c}
        \toprule
        Module & \# Params {[M]} & Acc@1 {[\%]}\\ \midrule
        ViT-S & 22 & 79.9 \\
        \textbf{(A)} & 21 & 80.1 \\
        \textbf{(B)} & 22 & 80.3 \\
        \textbf{(C)} & 29 & 80.6\\
        \midrule 
        $\gs$ (ours) & 22 & \textbf{80.8} \\
        $\gss$ (ours) & 22 & 80.4 \\
        \bottomrule
    \end{tabular}
    \vspace{-2mm}
    \caption{
    \textbf{Injection of uniform attention to ViT-S.}
    We inject uniform attention to $\msa$ \textbf{(A, B, C)} or $\mlp$ ($\gs$, $\gss$).
    ViT benefits from uniform attention.}
    \label{table:uniform_attention}
    \vspace{-1em}
\end{table}

\begin{figure}
        \centering
        \includegraphics[width=0.5\linewidth]{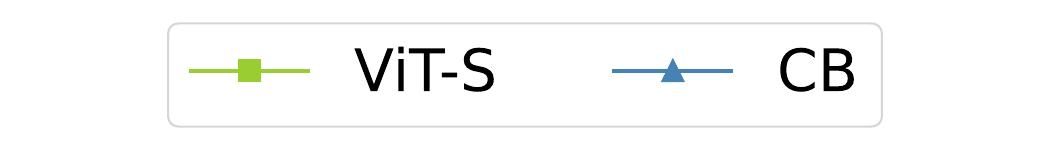}
        
        \begin{subfigure}[b]{0.49\linewidth}
            \centering
            \includegraphics[width=1.0\linewidth]{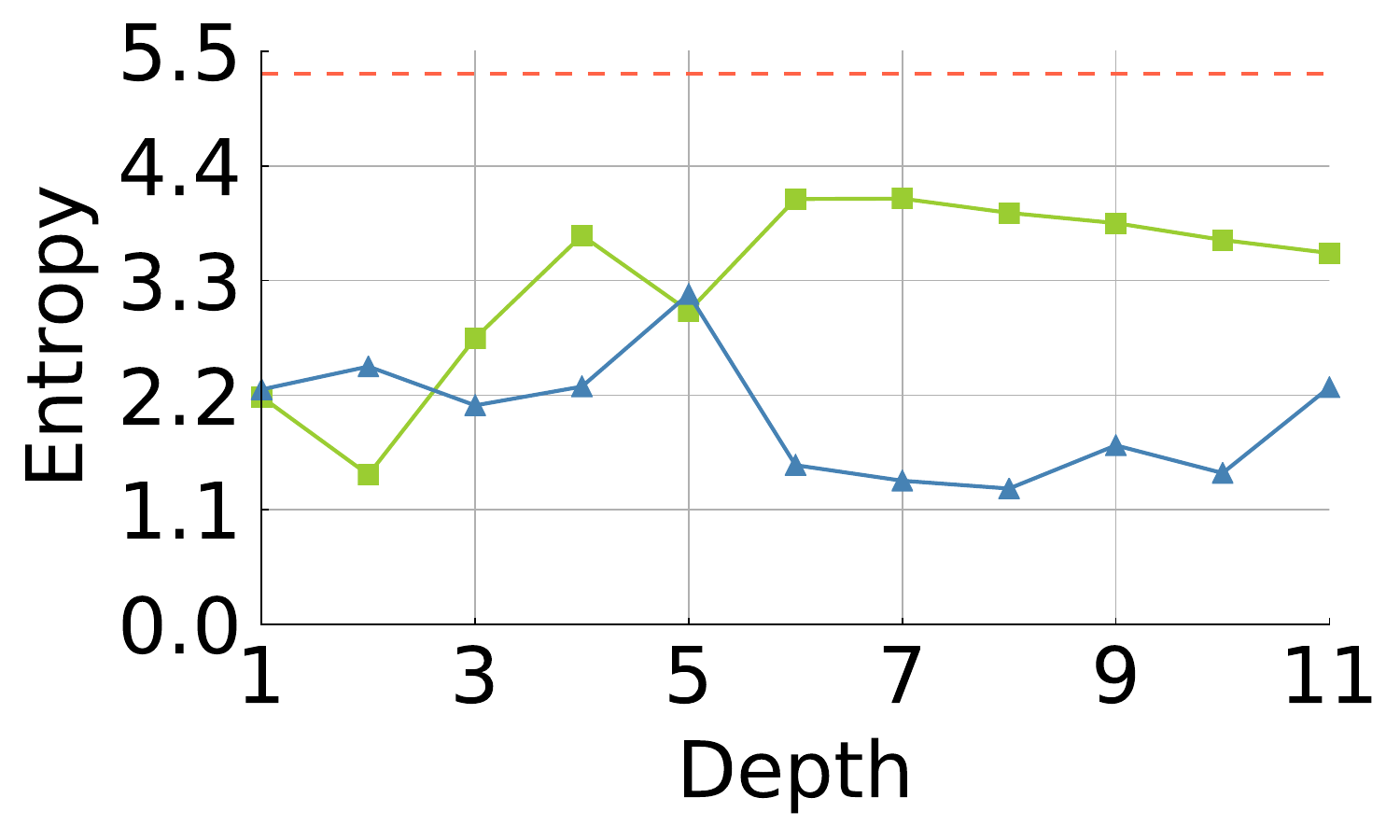}
            \vspace{-2em}
            \caption{Layer-wise entropy}
            \label{fig:entropya}
        \end{subfigure}
        \hfill
        \begin{subfigure}[b]{0.49\linewidth}
            \centering
            \includegraphics[width=1.0\linewidth]{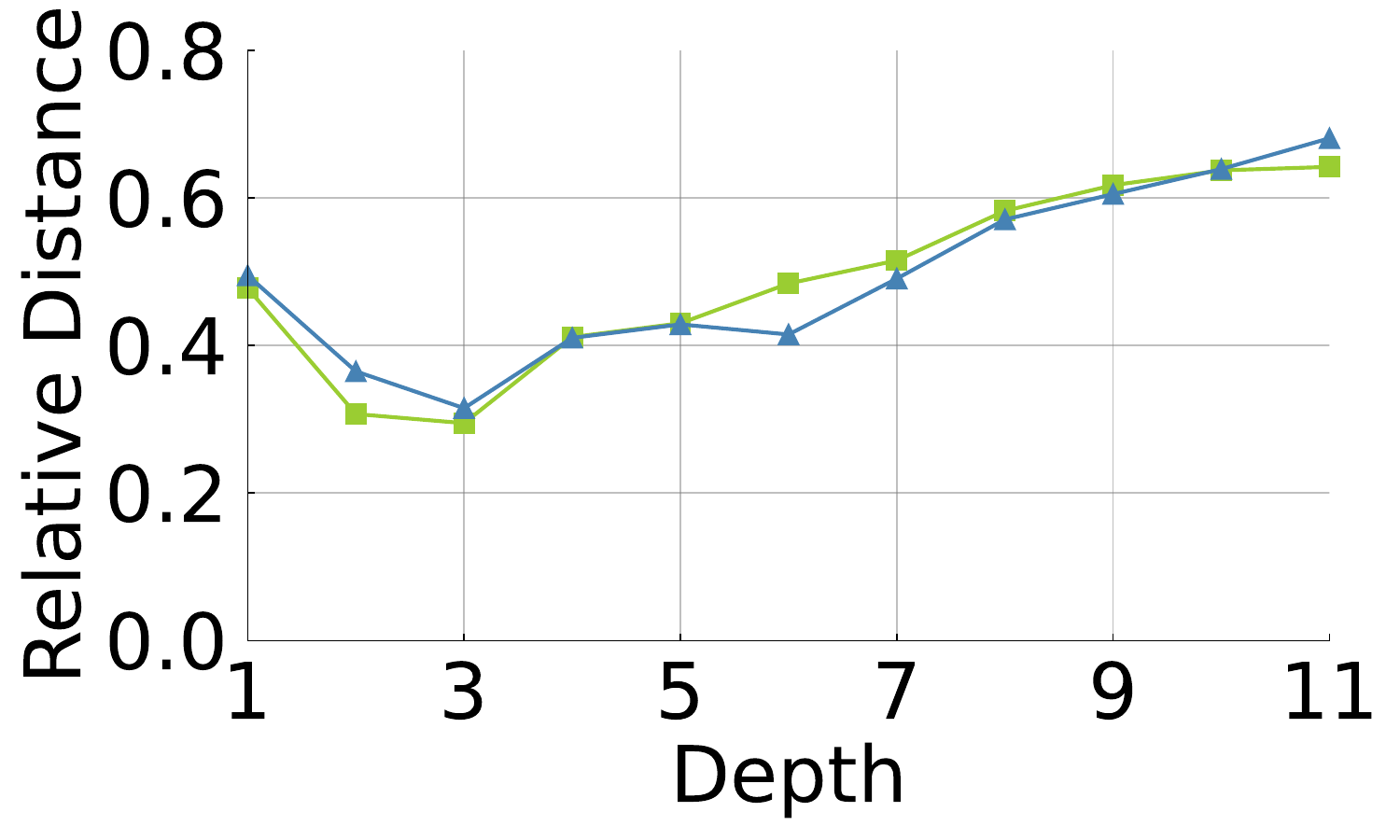}
            \vspace{-2em}
            \caption{Relative distance of attention}
            \label{fig:spa_dist}
        \end{subfigure} 
        \vspace{-.75em}
        \caption{\textbf{Attention entropy and relative distance.} We visualize the averaged entropies of the class token and the relative distance of spatial interactions across the layers. 
        $\gs$ changes the spatial interactions of attention and reserves the long-range dependency.
        }
        \label{fig:entropy}
        \vspace{-1em}
\end{figure}

\paragraph{Attention entropy according to the depth.}
We have observed in \Sref{sec:motivation} that the entropy of learned attention in ViT models tends to be high. 
From that, we have hypothesized that ViTs may benefit from an explicit injection of uniform attention.
We examine now whether our $\gs$ module lowers the burden of the self-attention to learn dense interactions.
We compare the entropy of the attention maps between ViT models with and without our $\gs$ module.
Figure~\ref{fig:entropya} shows layer-wise entropy values on ViT-S with and without our $\gs$ module.
The insertion of $\gs$ lowers the entropy values significantly, especially in deeper layers.
It seems that $\gs$ relaxes the representational burden for the $\msa$ block and lets $\msa$ focus on sparse interactions.

\paragraph{Relative distance according to the depth.}
We compute the relative distance of spatial interactions to see whether $\gs$ affects the range of spatial interactions.
We define the distance as follows: $\mathtt{dist}=\mathbb{E}_{i \neq j, \{i,j\} \in [1,N]}(a_{ij} || \bp_i - \bp_j ||_1)$, where $N$ is the number of spatial tokens, $a_{ij}$ is the weight of attention between $i$-th and $j$-th tokens, and $\bp_i$ is the normalized coordinate of $i$-th token.
We exclude the case of self-interaction to analyze interactions of other tokens.
As shown in \Fref{fig:spa_dist}, ViT-S and $\gs$ have a similar tendency.
Injecting the dense global interactions into ViT does not hurt the range of interactions.

\begin{figure}
    \small
    \centering
    \hspace{-3mm}
    \begin{minipage}[t]{.57\linewidth}
        \centering
        \includegraphics[width=.8\linewidth]{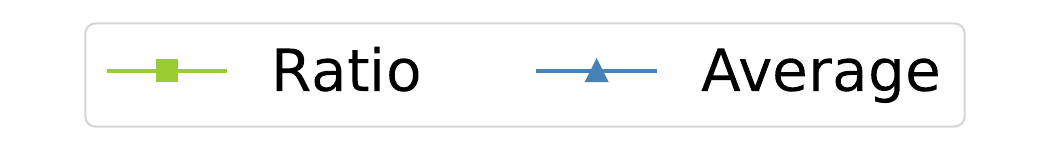}

        \includegraphics[width=0.9\linewidth]{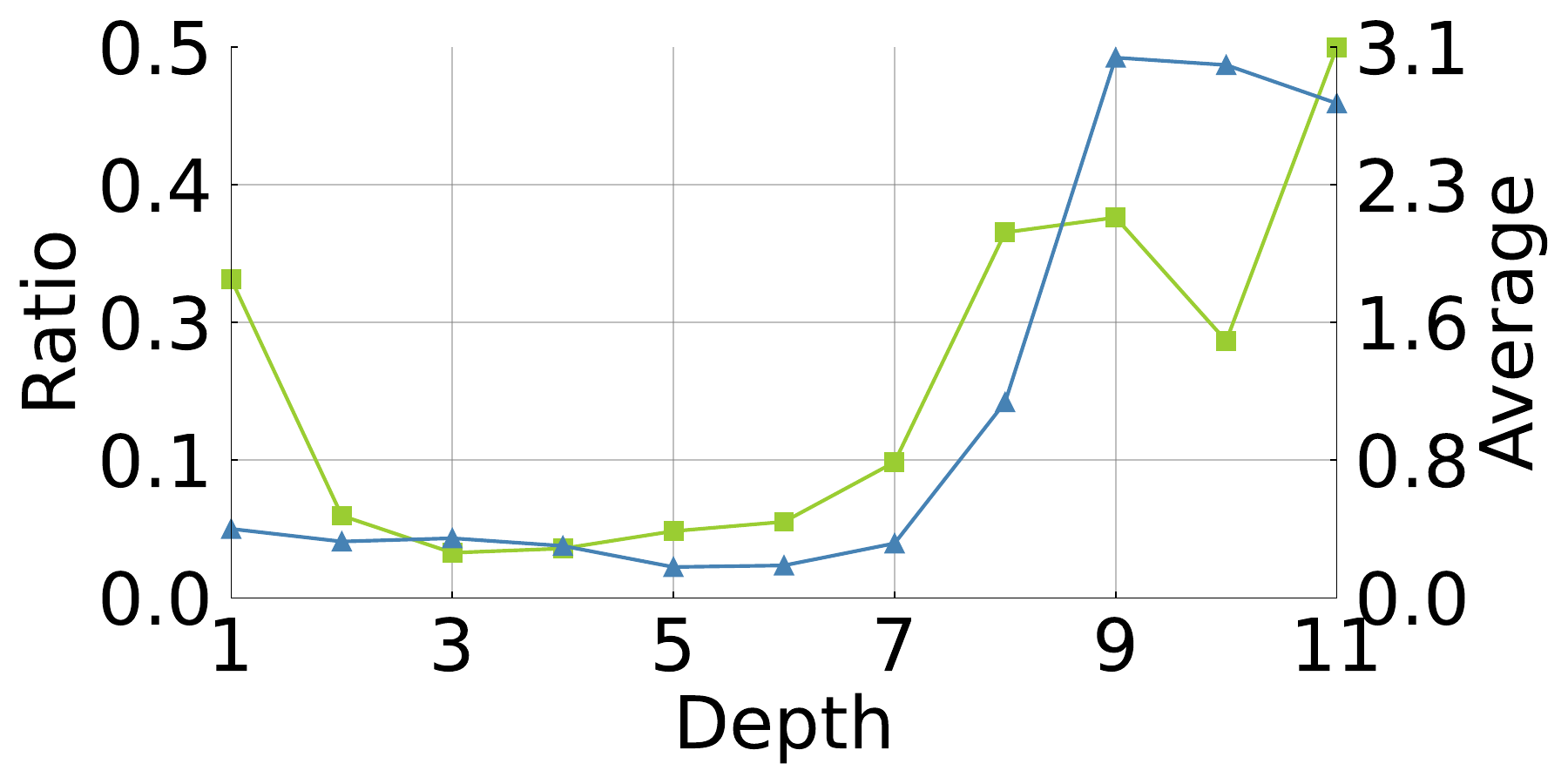}
        \captionsetup{width=0.9\linewidth}
        \captionof{figure}{\textbf{Analysis of dimension scaling.} 
        We plot values of the ratio and average of scaling weights across the layers.
        The high ratio and average indicate the preference for dense interactions.
        }
        \label{fig:a_scaling}
    \end{minipage}
    \hspace{-3mm}
    \begin{minipage}[t]{.47\linewidth}
        \centering
        \includegraphics[width=1.0\linewidth]{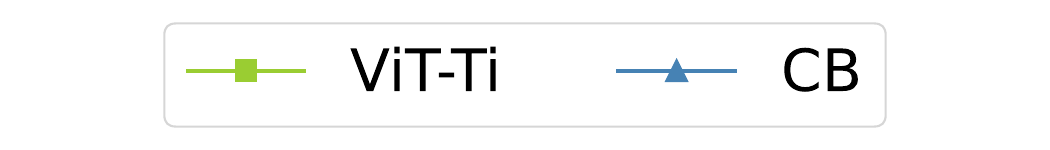}

        \includegraphics[width=.9\linewidth]{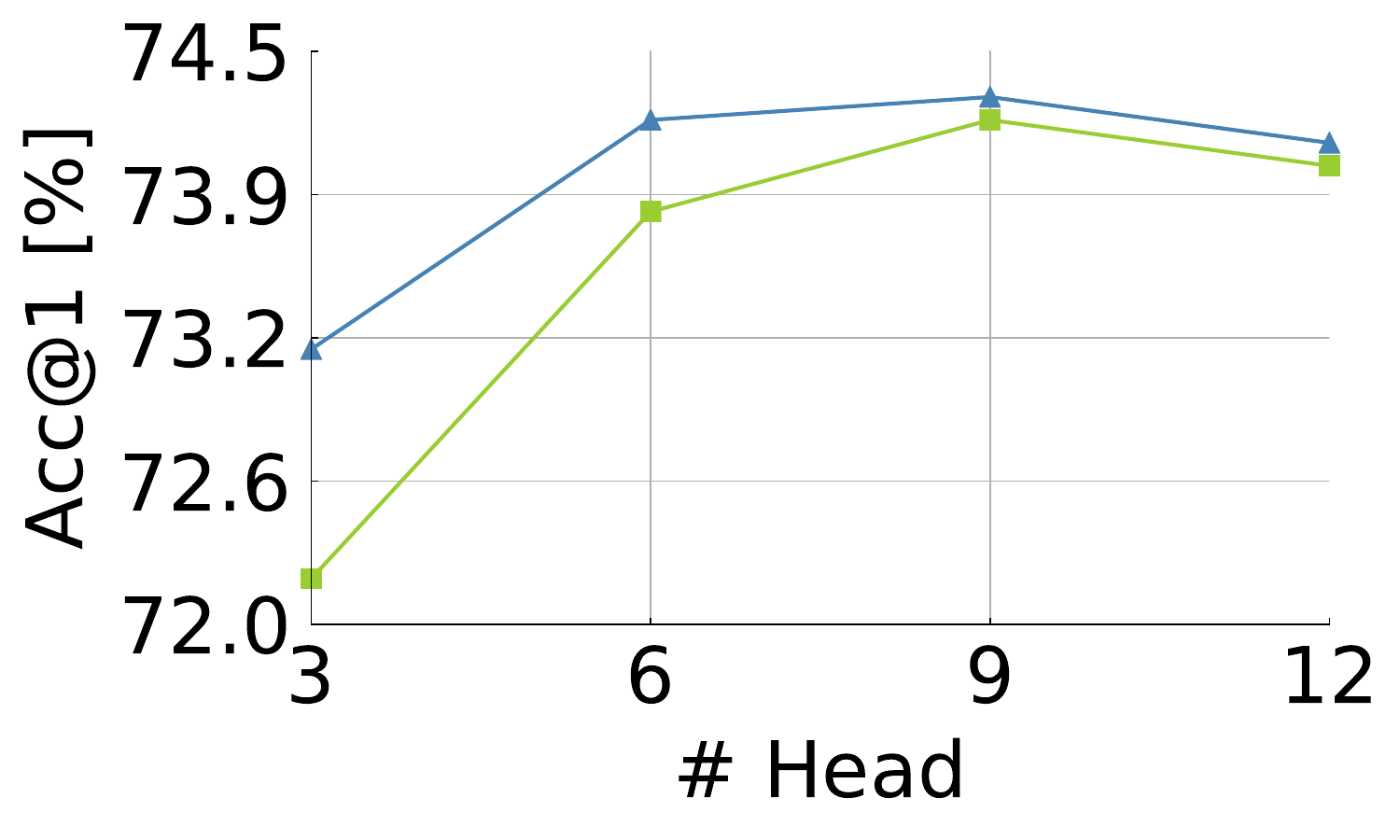}
        \captionsetup{width=0.9\linewidth}
        \captionof{figure}{\textbf{Accuracy vs. \# heads.} $\gs$ is effective with fewer heads where the lack of abundant spatial interactions may happen.}
        \label{fig:multi-head}
    \end{minipage}
\end{figure}

\paragraph{Analysis on dimension scaling.}
We analyze the magnitude of scaling weights $\lambda \in \bLambda$ in $\gss$ to identify the trend of the need for uniform attention according to depth.
We measure the ratio of the quantile of $90\%$ and $10\%$, $| \lambda_{0.1} | / | \lambda_{0.9} |$.
The ratio tells us how much high and low values of scaling weights are similar.
We also compute the average of scaling weights according to depth.
The average is related to the importance of uniform attention.
As shown in \Fref{fig:a_scaling}, the ratio and average increase along with the depth.
This indicates upper layers prefer dense interactions more than lower ones.
The result coincides with the above observation of entropy analysis, as shown in \Fref{fig:entropya}.

\begin{table}
    \centering
    \tabcolsep=0.3cm    
    \small
    \begin{tabular}{c c c c}\\
        \toprule  
        Model & No module & $\gs^\dagger$  & $\gs$ \\
        \midrule
        ViT-Ti & 72.2 & 73.2 & \textbf{73.4}\\
        ViT-S & 79.9 & 80.5 & \textbf{80.8} \\
        ViT-B & 81.8 & 82.0 & \textbf{82.1} \\
        \bottomrule
    \end{tabular}
    \vspace{-2mm}
    \caption{\textbf{ImageNet-1K performance of $\gs$.} 
    We denote $\gs^\dagger$ as $\gs$ applied to all layers.}
    \label{tab:upper}
    \vspace{-2mm}
\end{table}

\paragraph{Deeper Layers Need More Dense Interaction.}
As shown in \Fref{fig:entropya} and \Fref{fig:a_scaling}, we observe that ViTs prefer dense interactions in the deeper layers.
We compare infusing $\gs$ to all layers and upper layers.
We denote the insertion of all layers as $\gs^\dagger$.
As shown in Table~\ref{tab:upper}, $\gs$ achieves 1.2\%p, 0.9\%p, and 0.3\%p higher accuracy than the vanilla ViT-Ti/-S/-B, respectively.
$\gs$ also increases the top-1 accuracy further by 0.2\%p, 0.3\%p, and 0.1\%p compared to ViT-Ti/-S/-B with $\gs^\dagger$.
Inserting $\gs$ in deeper layers improves the performance further; thus, deeper layers benefit more from the dense interactions.

\paragraph{Accuracy according to the number of heads.}
$\msa$ can model abundant spatial interactions between tokens as the number of attention heads increases.
To examine the relationship between the number of heads and spatial interactions in $\msa$, we train ViT-Ti with and without $\gs$ by adjusting the number of heads of $\msa$.
As shown in \Fref{fig:multi-head}, the accuracy gap increases as the number of heads decreases.
Our proposed module is, therefore, more effective in a lower number of heads rather than the large number of heads.

\begin{table}
    \centering
    \small
    \hspace{-5.5mm}
    \begin{subtable}[h]{0.67\linewidth}
        \centering
        \tabcolsep=0.1cm
        \begin{tabular}{l c c}
             \toprule
             Extra resources & SE & $\gs$ \\
             \midrule
             {\footnotesize Extra parameters} & {\footnotesize Yes} & \textbf{\footnotesize No} \\
             {\footnotesize Computation costs} & {\footnotesize High} & \textbf{\footnotesize Low} \\
             {\footnotesize Implementation difficulty} & {\footnotesize Easy} & \textbf{\footnotesize 1 line} \\
             \bottomrule
        \end{tabular}
       \caption{\small{ Extra resources.}}
       \label{table:comp_se}
    \end{subtable}
    \hspace{-3mm}
    \begin{subtable}[h]{0.35\linewidth}
        \centering
        \tabcolsep=0.1cm
        \begin{tabular}{l c}
             \toprule
             Module & Acc@1 [\%] \\
             \midrule
             ViT-S & 79.9 \\
             + SE~\cite{hu2018squeeze} & 80.3\ \\
             + $\gs$ & \textbf{80.8} \\
             \bottomrule
        \end{tabular}
        \caption{\small {ImageNet-1K acc.}}
        \label{table:exp_se}
     \end{subtable}
     \vspace{-.75em}
     \caption{\textbf{Comparison with SE and $\gs$.} (a) Comparison in terms of the use of parameters, computation costs, and implementation difficulty.
     (b) Comparison of ImageNet-1K performance. Our $\gs$ contributes more to ViT-S compared with SE.}
     \label{tab:comp_se_}
     \vspace{-1em}
\end{table}
\paragraph{Comparison against SE.}
The SE module \cite{hu2018squeeze} shares a certain similarity to $\gs$: both are modular attachments to neural network architecture.
However, SE is designed to model the channel inter-dependency by exploiting pooling to construct a channel descriptor, two FC layers, and a sigmoid function.
See the comparison between $\gs$ and SE in \Tref{table:comp_se}.
Finally, we compare the performance of the models with SE and $\gs$. 
As shown in \Tref{table:exp_se}, $\gs$ and SE improve the accuracy by 0.9\%p and 0.4\%p, respectively.
Both modules improve the performance of ViT models, but the improvement is greater for $\gs$.

\paragraph{Conclusion.}
We observe that the global dense interaction enhances the performance of ViTs and diverts the role of $\msa$ to sparse interaction without reducing the distance of interaction.
It validates that the injection of useful interactions helps $\msa$ focus on other interactions.
We believe exploring other sophisticated explicit interactions will further benefit $\msa$.
In \Sref{sec:experiments}, we present the results of typical experiments based on our simple module.

\section{Experiments}\label{sec:experiments}
In \Sref{subsec:where}, we experiment with which location we put our module in. 
In Secs.~\ref{subsec:cls}-\ref{sec:obj_det}, we evaluate our modules on image classification, semantic segmentation, and object detection tasks.
\Sref{sec:sem_vis} provides the visualization of attention maps from ViT-S fine-tuned on segmentation task.
In \Sref{sec:robo}, we show results on the robustness benchmarks, including occlusion
and adversarial attack.
In Secs.~\ref{sec:vl} and \ref{sec:other_arch}, we evaluate our module on the vision-language Transformer for the Visual Question Answering task 
and on other architectures.

\begin{table}
    \centering
    \small
    \begin{subtable}[h]{1.0\linewidth}
        \centering
        \begin{tabular}{c c c c c c}
        
             \toprule
             \multirow{2}[2]{*}{Module} & \multicolumn{2}{c}{Position} & \multirow{2}[2]{*}{FLOPs {[G]}} & \multirow{2}[2]{*}{Acc@1 {[\%]}} \\\cmidrule(lr){2-3}
              & $\mlp$ & $\msa$ & & \\
             \midrule
             ViT-S & \xmark & \xmark & 4.6 & 79.9 \\ 
             \midrule
              \multirow{3}{*}{$\gs$} & \cmark & \xmark & 4.6 & \textbf{80.5} \\
              & \xmark & \cmark & 4.6 & 80.1 \\
               & \cmark & \cmark & 4.6 & 80.1 \\ \bottomrule
        \end{tabular}
        \vspace{-.25em}
       \caption{Position of $\gs$ to $\mlp$ and $\msa$.}
       \label{table:gcb_pos_blocks}
       \vspace{.25em}
    \end{subtable}
    \begin{subtable}[h]{1.0\linewidth}
        \centering
        \tabcolsep=0.1cm
        \begin{tabular}{c c c c c c c}
             \toprule
             \multirow{2}[2]{*}{Module} & \multicolumn{3}{c}{Position} & \multirow{2}[2]{*}{FLOPs {[G]}} & \multirow{2}[2]{*}{Acc@1 {[\%]}}\\ \cmidrule(lr){2-4}
              & $\Front$ & $\Mid$ & $\End$ & &  \\ \midrule
             ViT-S & \xmark & \xmark & \xmark & 4.6 & 79.9\\ \midrule
              \multirow{3}{*}{$\gs$} & \cmark & \xmark & \xmark & 4.6 & 79.9 \\
              & \xmark & \cmark & \xmark & 4.6 &  \textbf{80.5} \\
              & \xmark & \xmark & \cmark & 4.6 & \textbf{80.5} \\ 
              \bottomrule
        \end{tabular}
        \vspace{-.25em}
        \caption{Position of $\gs$ in $\mlp$.}
        \label{table:gcb_pos_mlp}
     \end{subtable}
     \vspace{-2mm}
     \caption{\textbf{Experiments with the position of $\gs$.} (a) ImageNet-1K performance when $\gs$ is inserted to either $\mlp$ and $\msa$.
     (b) ImageNet-1K performance when $\gs$ is placed at $\Front$, $\Mid$, or $\End$ in an $\mlp$ block.}
     \vspace{-4mm}
\end{table}

\begin{table*}
    \centering
    \small
    \begin{tabular}{l c c c c c c}
         \toprule
          Architecture & \# Params [M] & FLOPs {[G]} & Acc@1 {[\%]} & Acc@5 {[\%]} & IN-V2 {[\%]} & IN-ReaL [\%] \\
         \midrule
         ViT-Ti & 5.7 & 1.3 & 72.2 & 91.1 & 59.9 & 80.1\\
         $+ \ \gs$ & 5.7 &  1.3 & 73.4 & \textbf{91.9} & 61.3 & 81.0 \\
         $+ \ \gss$ & 5.7 & 1.3 & \textbf{73.5} & \textbf{91.9} & \textbf{61.4} & \textbf{81.2} \\
         \midrule
         ViT-S &  22.0 & 4.6 & 79.9 & 95.0 & 68.1 & 85.7 \\
         $+ \ \gs$ & 22.0 & 4.6 & \textbf{80.8} & \textbf{95.4} & \textbf{69.3} & \textbf{86.2} \\
         $+ \ \gss$ & 22.0 & 4.6 & 80.4 & 95.1 & 68.7 & 85.9 \\
         \midrule
         ViT-B\tablefootnote{We increase the warm-up epochs for learning stability in ViT-B.} & 86.6 & 17.6 & 81.8 & 95.6 & 70.5 & 86.7 \\
         $+ \ \gs$  & 86.6 & 17.6 & \textbf{82.1} & 95.7 & \textbf{71.1} & \textbf{86.9} \\
         $+ \ \gss$ & 86.6 & 17.6 & \textbf{82.1} & \textbf{95.8} & \textbf{71.1} & \textbf{86.9} \\
         \bottomrule
    \end{tabular}
    \vspace{-0.75em}
    \caption{\small \textbf{ImageNet-1K performance.} We train vision transformer architectures~\cite{vit, deit} with $\gs$ and $\gss$ and evaluate the accuracy on ImageNet-1K~\cite{deng2009imagenet}, ImageNet-V2~\cite{recht2019imagenet}, and ImageNet-ReaL~\cite{beyer2020we}. 
    \textbf{Bold} is the best number at each row. 
    Our module improves all the metrics incurring negligible extra computational costs.
    }
    \label{table:imnet}    
    \vspace{-1em}
\end{table*}
\subsection{Where to Insert CB in a ViT}\label{subsec:where}
We study the best location for $\gs$ with respect to the main blocks for ViT architectures: $\msa$ and $\mlp$.
We train ViT-S with our module positioned on $\mlp$, $\msa$, and both and validate on ImageNet-1K.
For simplicity, we infuse $\gs$ into all layers.
Note that this setting is different from the experiment in \Tref{table:uniform_attention}.
We place $\gs$ to the main block without complex adjustments.
As shown in \Tref{table:gcb_pos_blocks}, $\gs$ improves the performance regardless of blocks but achieves higher accuracy by 0.4\%p in an $\mlp$ block than either in an $\msa$ block or both.
It is notable, though, that adding $\gs$ increases the performance regardless of the location.
We have chosen $\mlp$ as the default location of our $\gs$ module for the rest of the paper.
This means that the self-attention and uniform attentions conduct their operation in $\msa$ and $\mlp$ alternately.
The alternation pattern considering the responsibility of modules can be found in prior work \cite{arnab2021vivit, prakash2021multi}.

Now, we study the best position of the $\gs$ module \emph{within} an $\mlp$ block, which consists of two fully-connected (FC) layers and the Gaussian Error Linear Unit (GELU) non-linear activation function \cite{hendrycks2016gaussian}.
Omitting the activation function for simplicity, 
we have three possible positions for $\gs$: $ \mathtt{<Front> - FC layer - <Mid> - FC layer - <End>}$.
We train ViT-S with $\gs$ located at $\Front$, $\Mid$, and $\End$, and validate on ImageNet-1K.
\Tref{table:gcb_pos_mlp} shows the performance; $\Mid$ and $\End$ increase accuracy by 0.6\%p compared to the vanilla ViT-S.
$\Mid$ demands four times larger computation costs than $\End$ because an $\mlp$ layer expands its channel dimensions four times rather than $\Front$ and $\End$.
We conclude that inserting $\gs$ at $\End$ of $\mlp$ tends to produce the best results overall.

Why is the improvement of the rear position larger than others? 
We conjecture that the gradient signal propagates to all parameters when $\gs$ is located at the $\End$ of $\mlp$ compared to being at the other places.
For simplicity, we assume a single layer composed of the $\msa$ and $\mlp$ blocks.
If $\gs$ is located at $\End$, the preceding weights in the $\msa$ and $\mlp$ block are updated by the gradient signals by uniform attention.
If $\gs$ is located at $\Front$, the subsequent weights in the corresponding $\mlp$ block cannot receive the gradient signals during training.

Why is the improvement of $\Mid$ and $\End$ similar?
There is no non-linear function (e.g., GELU) between $\Mid$ and $\End$ positions. 
Since uniform attention is the addition of a globally averaged token, the output is identical wherever $\gs$ is located at $\Mid$ and $\End$. 
Therefore, the accuracy of both positions is similar.

As a further study, we compare infusing $\gs$ to all layers or upper layers.
$\gs$ to upper layers achieves higher 
top-1 accuracy 
compared to $\gs$ to all layers in ViT-Ti/-S/-B.\footnote{The experiment can be found in Appendix.}
Inserting $\gs$ in deeper layers improves the performance further; thus, deeper layers benefit more from the dense interactions.

\subsection{Image Classification} \label{subsec:cls}
We train ViTs~\cite{vit} with our $\gs$ module on the ImageNet-1k training set and report accuracy on the validation set. We adopt strong regularizations following the DeiT \cite{deit}. 
We apply the random resized crop, random horizontal flip, Mixup \cite{zhang2018mixup}, CutMix \cite{yun2019cutmix}, random erasing \cite{zhong2020random}, repeated augmentations \cite{hoffer2020augment}, label-smoothing \cite{szegedy2016rethinking}, and stochastic depth \cite{huang2016deep}.
We use AdamW \cite{loshchilov2018decoupled} with betas of (0.9, 0.999), a learning rate of $10^{-3}{\cdot}\mathrm{(batch \ size)} / 1024$, and a weight decay of $0.05$.
The one-cycle cosine scheduling is used to decay the learning rate during the total epochs of 300.
We implement based on $\mathtt{PyTorch}$ \cite{NEURIPS2019_9015} and $\mathtt{timm}$ \cite{rw2019timm} on 8 V100 GPUs.
We use $\mathtt{torchprofile}$ library to count the number of FLOPs.
More details and additional experiments can be found in Appendix.

ViT-Ti/-S/-B~\cite{vit} with our modules trained on ImageNet-1K are further validated on ImageNet-V2~\cite{recht2019imagenet} and ImageNet-Real \cite{beyer2020we}.
\Tref{table:imnet} shows our modules $\gs$ and $\gss$ improve both precision and robustness of a model. 
$\gs$ does not add extra parameters, and $\gss$ increases only a few parameters; our modules demand negligible computation costs yet are effective for image classification.
The results signify our observations about the preference and learning difficulty of dense global attention and injecting dense attention explicitly are all valid.

\begin{table}[]
    \centering
    \small
    \tabcolsep=0.3cm
    \begin{tabular}{l c c c}
         \toprule
         \multirow{2}[2]{*}{Backbone} & \multirow{2}[2]{*}{\# Params [M]} & \multicolumn{2}{c}{mIoU [\%]}\\\cmidrule{3-4}
         & & \multicolumn{1}{c}{40K} & \multicolumn{1}{c}{160K}\\
         \midrule
         ViT-Ti & \multirow{3}{*}{34.1} & 35.5 & 38.9\\
         $+ \ \gs$ & & \textbf{36.5} & 39.0 \\
         $+ \ \gss$ & & 36.1 & \textbf{39.8} \\
         \midrule
         ViT-S & \multirow{3}{*}{53.5} & 41.5 & 43.3 \\
         $+ \ \gs$ & & \textbf{41.9} & \textbf{43.9} \\
         $+ \ \gss$ & & 41.6 & 43.1 \\
         \midrule
         ViT-B & \multirow{3}{*}{127.0} & 44.3 & 45.0 \\
         $+ \ \gs$ & & \textbf{45.1} & \textbf{45.6} \\
         $+ \ \gss$ & & 44.6 & 45.3 \\
         \bottomrule
    \end{tabular}
    \vspace{-0.75em}
    \caption{\textbf{ADE20K performance.} All models are based on UperNet~\cite{xiao2018unified}. Ours significantly improves the performance, and this is presumably because our module supplements global attention more to ViTs (like the atrous convolution~\cite{chen2017deeplab}).}
    \label{table:ade}
    \vspace{-1em}
\end{table}

\subsection{Semantic Segmentation} \label{subsec:seg}
We validate our method for semantic segmentation on the ADE20K dataset \cite{adk20k, adk20k_2} consisting of 20K training and 5K validation images.
For a fair comparison, we follow the protocol of XCiT \cite{el-nouby2021xcit} and Swin Transformer~\cite{swin}.
We adopt UperNet \cite{xiao2018unified} and train for 40K iterations or 160K for longer training.
Hyperparameters are the same as XCiT: the batch size of 16, AdamW with betas of (0.9, 0.999), the learning rate of $6\cdot10^{-5}$, weight decay of 0.01, and polynomial learning rate scheduling.
We set the head dimension as 192, 384, and 512 for ViT-Ti/-S/-B, respectively.
Table~\ref{table:ade} shows the results with 40K and 160K training settings.

We observe that ViT-Ti/-S/-B with $\gs$ increase mIoU by 1.0, 0.4, and 0.8 for 40K iterations and 0.1, 0.6, and 0.6 for 160K iterations, respectively.
Similarly, $\gss$ improves mIoU except for 160K iterations in ViT-S.
Infusing the context shows improvement in semantic segmentation; the performance improvement of ViT-B is not marginal, especially.
The result would be related to the prior work \cite{chen2017deeplab, zhao2017pyramid}, 
which introduces the global context by atrous convolution and pyramid module.
$\gs$ not only performs the dense interactions across tokens, which the original self-attention is hard to learn, but also supplies the global context.

\begin{table}
    \centering
    \small
    \begin{tabular}{c c c c | c c c}
        \toprule
        Model & AP$^b$ & AP$^b_{50}$ & AP$^b_{75}$ & AP$^m$ & AP$^m_{50}$ & AP$^m_{75}$\\
        \midrule
        ViT-Ti & 34.8 & 57.4 & 36.5 & 32.5 & 54.3 & 33.7\\
        + $\gs$ & \textbf{35.1} & \textbf{57.9} & \textbf{36.8} & \textbf{32.8} & \textbf{54.4} & \textbf{34.2}\\
        \bottomrule
    \end{tabular}
    \vspace{-0.75em}
    \caption{\textbf{COCO object detection and instance segmentation performance.} 
    We finetune ViT-Ti on the COCO dataset for 12 epochs (1$\times$ schedule).}
    \label{tab:coco}
    \vspace{-1em}
\end{table}

\subsection{Object detection}\label{sec:obj_det}
We fine-tune the pre-trained ViT-Ti on the COCO dataset and evaluate the performance of object detection and instance segmentation in \Tref{tab:coco}.
COCO consists of 118K training and 5K validation images with 80 categories.
We follow the protocol of XCiT \cite{el-nouby2021xcit} and Swin Transformer~\cite{swin}.
We adopt Mask R-CNN with FPN and train models for 12 epochs ($1\times$ schedule) using AdamW with learning rate $10^{-4} {\cdot} \tfrac{\text{batch size}}{16}$ and weight decay 0.05.
We do not apply $\gs$ to a block where features are feed-forwarded to FPN.
Ours consistently improves performance.
In object detection, $\gs$ improves 0.3, 0.5, and 0.3 in AP$^b$, AP$^b_{50}$, and AP$^b_{75}$, respectively.
In instance segmentation, $\gs$ improves 0.3, 0.1, and 0.5 in AP$^m$, AP$^m_{50}$, and AP$^m_{75}$, respectively.

\begin{figure*}
    \centering
    \includegraphics[width=0.7\linewidth]{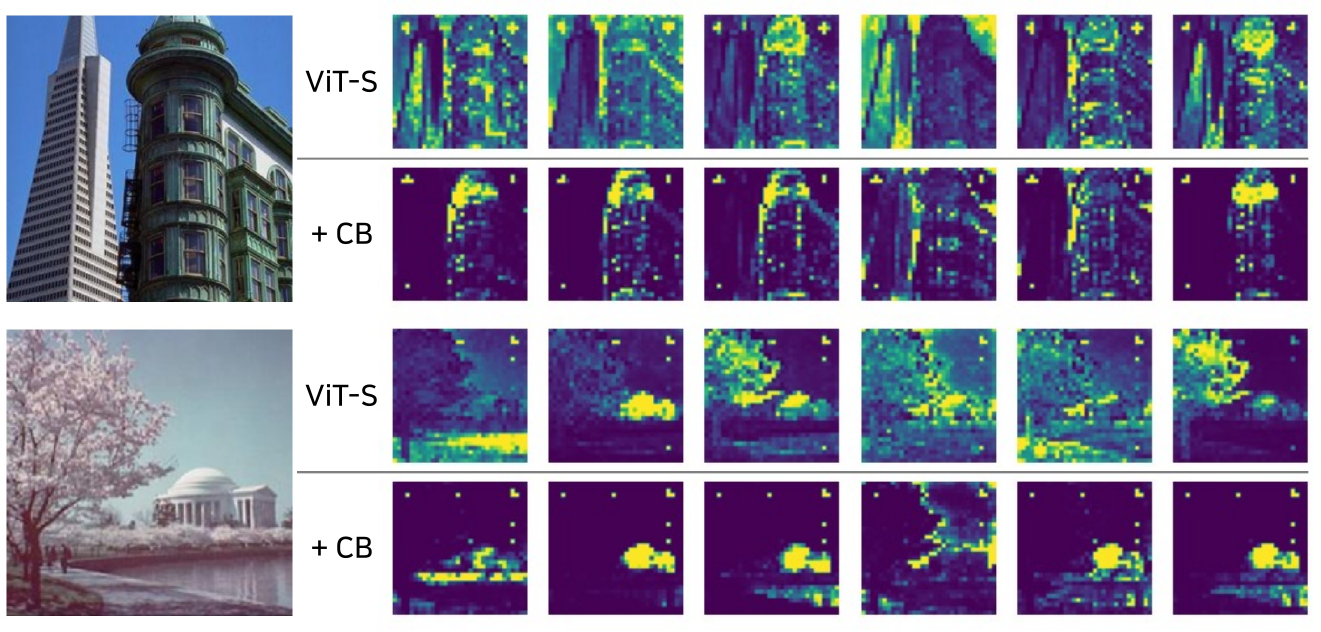}
    \vspace{-1em}
    \caption{\textbf{Visualization of attention maps.}
    Using ViT-S fine-tuned on ADE20K, we visualize the attention maps of the last layers of heads. 
    The first row of each image corresponds with ViT-S, and the second row does ViT-S with $\gs$.
    We can observe that $\gs$ reduces the dense aggregation of self-attention.
    By infusing uniform attention, $\msa$ aggregates more informative signals, such as objects.
    }
    \label{fig:attn_vis}
    \vspace{-1em}
\end{figure*}
\subsection{Segmentation Attention Visualization}\label{sec:sem_vis}
We visualize the attention maps to understand how $\gs$ changes the interactions of $\msa$ rather than entropies.
We use the pre-trained ViT-S on ADE20K to extract the attention maps.
The visualized attentions are extracted from the last layers before Feature Pyramid Network (FPN).
See \Fref{fig:attn_vis} for a comparison.
We apply the same thresholding and min-max normalization in visualization for a fair comparison.
ViT-S without $\gs$ needs dense aggregations more than ViT-S with $\gs$.
The visualization also validates that $\gs$ takes over the dense aggregations from the original self-attention.
This implies that $\gs$ splits the burden of self-attention.

\begin{table}[]
    \centering
    \small
    \tabcolsep=0.2cm
    \begin{tabular}{l c c c}
         \toprule
          Architecture & Occ [\%] & ImageNet-A [\%] & FGSM [\%]\\
         \midrule
         ViT-S & 73.0 & 19.0 & 27.2 \\
         $+ \ \gss$ & 73.7 & 19.1 & 27.8 \\
         $+ \ \gs$ & \textbf{74.0} & \textbf{21.2} & \textbf{32.3} \\
         \bottomrule
    \end{tabular}
    \vspace{-0.75em}
    \caption{\textbf{Robustness evaluation.} We evaluate ViT-S with $\gs$ and $\gss$ on center occlusion (Occ), ImageNet-A, and fast sign gradient method (FGSM) attack. Ours shows improved robustness across the board against ViT-S.}
    \label{table:robo}
    \vspace{-1em}
\end{table}

\subsection{Evaluating Model Robustness} \label{sec:robo}
We evaluate the robustness of $\gs$ and $\gss$ with respect to center occlusion (Occ), ImageNet-A \cite{hendrycks2021nae}, and an adversarial attack \cite{goodfellow2014explaining}.
For Occ, we zero-mask the center $112 \times 112$ patches of every validation image.
ImageNet-A is the collection of challenging test images that an ensemble of ResNet50s has failed to recognize.
We employ the fast sign gradient method (FGSM~\cite{goodfellow2014explaining}) for the adversarial attack.
Table~\ref{table:robo} shows the results of the robustness benchmark.
$\gss$ increases by 0.7, 0.1, and 0.6 of Occ, ImageNet-A, and FGSM, respectively.
$\gs$ does 1.0, 2.2, and 5.1, respectively.

\begin{table}
    \small
    \centering
    \resizebox{1.0\linewidth}{!}{
    \begin{tabular}{c c c}
         \toprule
         Noise Type & ViT-S & $\gs$ \\
         \midrule
         Nothing & 43.3 & 43.9 \\ 
         Shot Noise & 40.22 $\pm$ 0.15 & 41.09 $\pm$ 0.09 \\
         Gaussian Noise (sigma=5.0) & 42.55 $\pm$ 0.08 & 43.44 $\pm$ 0.08 \\
         Gaussian Noise (sigma=10.0) & 40.22 $\pm$ 0.07 & 41.07 $\pm$ 0.06 \\
         Gaussian Blur (sigma=1.0) & 42.29 & 43.26 \\
         Gaussian Blur (sigma=2.0) & 40.83 & 41.44 \\
         \bottomrule
    \end{tabular}}
    \vspace{-0.75em}    
    \caption{\textbf{Robustness evaluation on ADE20K with input perturbations.} We evaluate ViT-S with and without $\gs$ on shot noise, Gaussian noise, and Gaussian blur. 
    The performance gap of perturbations between ViT-S and $\gs$ is larger than the one of nothing. 
    It shows that $\gs$ improves robustness.
    We run the experiments on random noise five times and report a mean with a confidence interval of 95\%.}
    \label{table:robust_ade20k}
    \vspace{-1em}    
\end{table}
We also evaluate the robustness on ADE20K using input perturbations~\cite{hendrycks2019benchmarking}, \eg, shot noise, Gaussian noise, and Gaussian blur. 
We run the experiments by five times on random noise and report the mean and confidence interval of 95\%. \Tref{table:robust_ade20k} shows the performance of mIoU.
The performance gap of ViT-S with and without $\gs$ increases from 0.6 up to 0.97. 
This shows that our 1 line of code can improve the ViT models’ robustness against input perturbation in the semantic segmentation task.

\subsection{Vision-Language Transformer} \label{sec:vl}
 \setlength{\intextsep}{0.5em}%
\setlength{\columnsep}{.5em}%
\begin{wraptable}{r}{.44\linewidth}
    \centering
    \small
    \vspace{-.5em}
    \tabcolsep=1mm
    \begin{tabular}{l c}
         \toprule
          Architecture & Acc [\%] \\
         \midrule
         ViLT~\cite{vilt} & 71.28 \\
         $+ \ \gss$ (Image) & 71.44 \\
         $+ \ \gss$ (Text) & \textbf{71.46} \\
         $+ \ \gss$ (Both) & 71.42 \\
         \bottomrule
    \end{tabular}
    \caption{\textbf{Vision language transformer results on VQAv2}. We fine-tune ViLT with $\gss$ on tokens of image, text, and both. }
    \label{table:vl}
\end{wraptable}

Transformer becomes the standard architecture for multi-modal learning because of the succinct structure. 
For example, Transformer employs modality-specific linear projection~\cite{vilt, hu2021unit, herzig2020tapas, akbari2021vatt}.
We evaluate our module on the Vision-Language Transformer, ViLT~\cite{vilt}.
We fine-tune the  pre-trained ViLT on VQAv2~\cite{balanced_vqa_v2} using the official code. 
Table~\ref{table:vl} shows the results of the performance. 
We first reproduce the baseline and reach the reported number (71.26). 
Our module is applied to the image, text, and both tokens, and in all cases, it improves the accuracy by 0.16, 0.18, and 0.14, respectively, compared with the baseline accuracy.

\begin{table}
    \centering
    \small
    \tabcolsep=0.15cm
    \begin{tabular}{c c c c c}
         \toprule
          \multirow{2}{*}{Architecture} & \multirow{2}{*}{\# Params [M]}  & \multirow{2}{*}{FLOPS {[G]}} & \multirow{2}{*}{Acc@1 {[\%]}} & \\
          & & & \\
         \midrule
         PiT-B~\cite{pit} & 73.8 & 12.4 & 82.0 \\
         $+ \ \gs$ & 73.8 & 12.4 & \textbf{82.6} \\
         \midrule
         Mixer-S/16~\cite{mlp_mixer} & 18.5 & 3.8 & 74.3 \\
         $+ \ \gs$ &18.5  & 3.8 & \textbf{74.9} \\
         \bottomrule
    \end{tabular}
    \vspace{-.75em}
    \caption{\textbf{ImageNet-1K performance on other architectures.}
    Ours also improves the performance in other models of Transformer and MLP.
    }
    \label{table:abla_feedforward}
    \vspace{-1em}
\end{table}
\subsection{Other Architectures}\label{sec:other_arch}
We evaluate $\gs$ on PiT \cite{pit} and Mixer \cite{mlp_mixer}.
PiT-B is the variant of the original Vision Transformer introducing spatial dimension reduction.
Mixer is pioneering work of the feed-forward architectures \cite{mlp_mixer, touvron2021resmlp}, mainly consisting of FC layers.
The structure of feed-forward architecture follows ViT except for $\msa$.
Spatial interactions of feed-forward are done through transposing visual data followed by an FC layer.
We insert our module at $\mlp$ in PiT and Mixer~\cite{mlp_mixer}.
For a fair comparison, we reproduce the baseline Mixer-S/16 with the DeiT training regime~\cite{deit} and train ours with the same one. 
Our module increases the performance of those architectures, as shown in \Tref{table:abla_feedforward}.

\section{Conclusion}\label{sec:conclusion}

We look closer at the spatial interactions in ViTs, especially in terms of density.
We have been motivated by the preliminary exploration and observations that suggest ViT models prefer dense interactions.
We also show that, at least from the optimization point of view, uniform attention is perhaps the most challenging attention for softmax-based attention to learn. 
The preference and optimization difficulty of learning dense interactions are not aligned.
It leads us to introduce further dense interactions manually by a simple module: Context Broadcasting ($\gs$).
Inserted at intermediate layers of ViT models, $\gs$ adds the averaged token to tokens.
Additionally, we propose a dimension scaling version of $\gs$, called $\gss$, to infuse the dense interactions selectively.
It turns out that our simple module improves the ViT performances across
various benchmarks, including image classification, semantic segmentation, and visual-language tasks.
$\gs$ only takes 1 line of code, a few more FLOPs, and zero parameters using this module. 
We hope that our module will further improve your ViT models and that our observations provide insights
for modeling the token interactions of ViTs.

Our work introduces the simplest form of dense interaction that complement self-attention.
One may propose a more sophisticated and effective module
that makes self-attention focus on the crucial interactions that should be only dealt with by self-attention, intractable otherwise.
We believe that this would be an exciting research direction.

\vspace{2mm}
\paragraph{Acknowledgment}
N.~Hyeon-Woo, K.~Y.-J. and T.-H.~Oh were partly supported by Institute of Information $\&$ communications Technology Planning $\&$ Evaluation (IITP) grant funded by the Korea government(MSIT) (No.2022-0-00124, Development of Artificial Intelligence Technology for Self-Improving Competency-Aware Learning Capabilities; No.2022-0-00290, Visual Intelligence for Space-Time Understanding and Generation based on Multi-layered Visual Common Sense; No. 2020-0-00004, Development of Previsional Intelligence based on Long-term Visual Memory Network).

\newpage
{\small
\bibliographystyle{ieee_fullname}

\begin{thebibliography}{10}\itemsep=-1pt

\bibitem{tensorflow}
Mart{\'\i}n Abadi, Paul Barham, Jianmin Chen, Zhifeng Chen, Andy Davis, Jeffrey Dean, Matthieu Devin, Sanjay Ghemawat, Geoffrey Irving, Michael Isard, et~al.
\newblock Tensorflow: A system for large-scale machine learning.
\newblock In {\em USENIX Symposium on Operating Systems Design and Implementation (OSDI)}, 2016.

\bibitem{akbari2021vatt}
Hassan Akbari, Liangzhe Yuan, Rui Qian, Wei-Hong Chuang, Shih-Fu Chang, Yin Cui, and Boqing Gong.
\newblock Vatt: Transformers for multimodal self-supervised learning from raw video, audio and text.
\newblock {\em NeurIPS}, 2021.

\bibitem{VQA}
Stanislaw Antol, Aishwarya Agrawal, Jiasen Lu, Margaret Mitchell, Dhruv Batra, C.~Lawrence Zitnick, and Devi Parikh.
\newblock {VQA}: {V}isual {Q}uestion {A}nswering.
\newblock In {\em ICCV}, 2015.

\bibitem{arnab2021vivit}
Anurag Arnab, Mostafa Dehghani, Georg Heigold, Chen Sun, Mario Lu{\v{c}}i{\'c}, and Cordelia Schmid.
\newblock Vivit: A video vision transformer.
\newblock In {\em ICCV}, 2021.

\bibitem{beyer2020we}
Lucas Beyer, Olivier~J H{\'e}naff, Alexander Kolesnikov, Xiaohua Zhai, and A{\"a}ron van~den Oord.
\newblock Are we done with imagenet?
\newblock {\em arXiv preprint arXiv:2006.07159}, 2020.

\bibitem{jax}
James Bradbury, Roy Frostig, Peter Hawkins, Matthew~James Johnson, Chris Leary, Dougal Maclaurin, George Necula, Adam Paszke, Jake Vander{P}las, Skye Wanderman-{M}ilne, and Qiao Zhang.
\newblock {JAX}: composable transformations of {P}ython+{N}um{P}y programs, 2018.

\bibitem{cao2022gcnet}
Yue Cao, Jiarui Xu, Stephen Lin, Fangyun Wei, and Han Hu.
\newblock Global context networks.
\newblock {\em IEEE TPAMI}, to appear.

\bibitem{carion2020end}
Nicolas Carion, Francisco Massa, Gabriel Synnaeve, Nicolas Usunier, Alexander Kirillov, and Sergey Zagoruyko.
\newblock End-to-end object detection with transformers.
\newblock In {\em ECCV}, 2020.

\bibitem{chen2017deeplab}
Liang-Chieh Chen, George Papandreou, Iasonas Kokkinos, Kevin Murphy, and Alan~L Yuille.
\newblock Deeplab: Semantic image segmentation with deep convolutional nets, atrous convolution, and fully connected crfs.
\newblock {\em IEEE TPAMI}, 40(4):834--848, 2017.

\bibitem{Cordonnier2020On}
Jean-Baptiste Cordonnier, Andreas Loukas, and Martin Jaggi.
\newblock On the relationship between self-attention and convolutional layers.
\newblock In {\em ICLR}, 2020.

\bibitem{dai2017deformable}
Jifeng Dai, Haozhi Qi, Yuwen Xiong, Yi Li, Guodong Zhang, Han Hu, and Yichen Wei.
\newblock Deformable convolutional networks.
\newblock In {\em ICCV}, 2017.

\bibitem{deng2009imagenet}
Jia Deng, Wei Dong, Richard Socher, Li-Jia Li, Kai Li, and Li Fei-Fei.
\newblock Imagenet: A large-scale hierarchical image database.
\newblock In {\em CVPR}, 2009.

\bibitem{vit}
Alexey Dosovitskiy, Lucas Beyer, Alexander Kolesnikov, Dirk Weissenborn, Xiaohua Zhai, Thomas Unterthiner, Mostafa Dehghani, Matthias Minderer, Georg Heigold, Sylvain Gelly, Jakob Uszkoreit, and Neil Houlsby.
\newblock An image is worth 16x16 words: Transformers for image recognition at scale.
\newblock In {\em ICLR}, 2021.

\bibitem{el-nouby2021xcit}
Alaaeldin El-Nouby, Hugo Touvron, Mathilde Caron, Piotr Bojanowski, Matthijs Douze, Armand Joulin, Ivan Laptev, Natalia Neverova, Gabriel Synnaeve, Jakob Verbeek, and Herve Jegou.
\newblock {XC}it: Cross-covariance image transformers.
\newblock In {\em NeurIPS}, 2021.

\bibitem{girdhar2022omnivore}
Rohit Girdhar, Mannat Singh, Nikhila Ravi, Laurens van~der Maaten, Armand Joulin, and Ishan Misra.
\newblock Omnivore: A single model for many visual modalities.
\newblock In {\em CVPR}, 2022.

\bibitem{goodfellow2014explaining}
Ian~J Goodfellow, Jonathon Shlens, and Christian Szegedy.
\newblock Explaining and harnessing adversarial examples.
\newblock {\em ICLR}, 2014.

\bibitem{balanced_vqa_v2}
Yash Goyal, Tejas Khot, Douglas Summers{-}Stay, Dhruv Batra, and Devi Parikh.
\newblock Making the {V} in {VQA} matter: Elevating the role of image understanding in {V}isual {Q}uestion {A}nswering.
\newblock In {\em CVPR}, 2017.

\bibitem{he2021pruning}
Haoyu He, Jing Liu, Zizheng Pan, Jianfei Cai, Jing Zhang, Dacheng Tao, and Bohan Zhuang.
\newblock Pruning self-attentions into convolutional layers in single path.
\newblock {\em arXiv preprint arXiv:2111.11802}, 2021.

\bibitem{resnet}
Kaiming He, Xiangyu Zhang, Shaoqing Ren, and Jian Sun.
\newblock Deep residual learning for image recognition.
\newblock In {\em CVPR}, 2016.

\bibitem{hendrycks2019benchmarking}
Dan Hendrycks and Thomas Dietterich.
\newblock Benchmarking neural network robustness to common corruptions and perturbations.
\newblock {\em arXiv preprint arXiv:1903.12261}, 2019.

\bibitem{hendrycks2016gaussian}
Dan Hendrycks and Kevin Gimpel.
\newblock Gaussian error linear units (gelus).
\newblock {\em arXiv preprint arXiv:1606.08415}, 2016.

\bibitem{hendrycks2021nae}
Dan Hendrycks, Kevin Zhao, Steven Basart, Jacob Steinhardt, and Dawn Song.
\newblock Natural adversarial examples.
\newblock {\em CVPR}, 2021.

\bibitem{pit}
Byeongho Heo, Sangdoo Yun, Dongyoon Han, Sanghyuk Chun, Junsuk Choe, and Seong~Joon Oh.
\newblock Rethinking spatial dimensions of vision transformers.
\newblock In {\em ICCV}, 2021.

\bibitem{herzig2020tapas}
Jonathan Herzig, Pawe{\l}~Krzysztof Nowak, Thomas M{\"u}ller, Francesco Piccinno, and Julian~Martin Eisenschlos.
\newblock Tapas: Weakly supervised table parsing via pre-training.
\newblock {\em arXiv preprint arXiv:2004.02349}, 2020.

\bibitem{hoffer2020augment}
Elad Hoffer, Tal Ben-Nun, Itay Hubara, Niv Giladi, Torsten Hoefler, and Daniel Soudry.
\newblock Augment your batch: Improving generalization through instance repetition.
\newblock In {\em CVPR}, 2020.

\bibitem{hu2018squeeze}
Jie Hu, Li Shen, and Gang Sun.
\newblock Squeeze-and-excitation networks.
\newblock In {\em CVPR}, 2018.

\bibitem{hu2021unit}
Ronghang Hu and Amanpreet Singh.
\newblock Unit: Multimodal multitask learning with a unified transformer.
\newblock In {\em ICCV}, 2021.

\bibitem{huang2017densely}
Gao Huang, Zhuang Liu, Laurens Van Der~Maaten, and Kilian~Q Weinberger.
\newblock Densely connected convolutional networks.
\newblock In {\em CVPR}, 2017.

\bibitem{huang2016deep}
Gao Huang, Yu Sun, Zhuang Liu, Daniel Sedra, and Kilian~Q Weinberger.
\newblock Deep networks with stochastic depth.
\newblock In {\em ECCV}, 2016.

\bibitem{huang2019ccnet}
Zilong Huang, Xinggang Wang, Lichao Huang, Chang Huang, Yunchao Wei, and Wenyu Liu.
\newblock Ccnet: Criss-cross attention for semantic segmentation.
\newblock In {\em ICCV}, 2019.

\bibitem{jiyeon2023mindvps}
Ji-Yeon Kim, Hyun-Bin Oh, Dahun Kim, and Tae-Hyun Oh.
\newblock Mindvps: Minimal model for depth-aware video panoptic segmentation.
\newblock In {\em IEEE Conference on Computer Vision and Pattern Recognition Workshops (CVPRW)}, 2023.

\bibitem{vilt}
Wonjae Kim, Bokyung Son, and Ildoo Kim.
\newblock Vilt: Vision-and-language transformer without convolution or region supervision.
\newblock In {\em ICML}, 2021.

\bibitem{lin2021end}
Kevin Lin, Lijuan Wang, and Zicheng Liu.
\newblock End-to-end human pose and mesh reconstruction with transformers.
\newblock In {\em CVPR}, 2021.

\bibitem{liu2015parsenet}
Wei Liu, Andrew Rabinovich, and Alexander~C Berg.
\newblock Parsenet: Looking wider to see better.
\newblock {\em arXiv preprint arXiv:1506.04579}, 2015.

\bibitem{swin}
Ze Liu, Yutong Lin, Yue Cao, Han Hu, Yixuan Wei, Zheng Zhang, Stephen Lin, and Baining Guo.
\newblock Swin transformer: Hierarchical vision transformer using shifted windows.
\newblock In {\em ICCV}, 2021.

\bibitem{loshchilov2018decoupled}
Ilya Loshchilov and Frank Hutter.
\newblock Decoupled weight decay regularization.
\newblock In {\em ICLR}, 2019.

\bibitem{luo2016understanding}
Wenjie Luo, Yujia Li, Raquel Urtasun, and Richard Zemel.
\newblock Understanding the effective receptive field in deep convolutional neural networks.
\newblock {\em NeurIPS}, 2016.

\bibitem{ma2022close}
Xu Ma, Huan Wang, Can Qin, Kunpeng Li, Xingchen Zhao, Jie Fu, and Yun Fu.
\newblock A close look at spatial modeling: From attention to convolution.
\newblock {\em arXiv preprint arXiv:2212.12552}, 2022.

\bibitem{mao2021transformer}
Yuxin Mao, Jing Zhang, Zhexiong Wan, Yuchao Dai, Aixuan Li, Yunqiu Lv, Xinyu Tian, Deng-Ping Fan, and Nick Barnes.
\newblock Transformer transforms salient object detection and camouflaged object detection.
\newblock {\em arXiv preprint arXiv:2104.10127}, 2021.

\bibitem{NEURIPS2021_c404a5ad}
Muhammad~Muzammal Naseer, Kanchana Ranasinghe, Salman~H Khan, Munawar Hayat, Fahad Shahbaz~Khan, and Ming-Hsuan Yang.
\newblock Intriguing properties of vision transformers.
\newblock In {\em NeurIPS}, 2021.

\bibitem{park2022how}
Namuk Park and Songkuk Kim.
\newblock How do vision transformers work?
\newblock In {\em ICLR}, 2022.

\bibitem{NEURIPS2019_9015}
Adam Paszke, Sam Gross, Francisco Massa, Adam Lerer, James Bradbury, Gregory Chanan, Trevor Killeen, Zeming Lin, Natalia Gimelshein, Luca Antiga, Alban Desmaison, Andreas Kopf, Edward Yang, Zachary DeVito, Martin Raison, Alykhan Tejani, Sasank Chilamkurthy, Benoit Steiner, Lu Fang, Junjie Bai, and Soumith Chintala.
\newblock Pytorch: An imperative style, high-performance deep learning library.
\newblock In {\em NeurIPS}, 2019.

\bibitem{prakash2021multi}
Aditya Prakash, Kashyap Chitta, and Andreas Geiger.
\newblock Multi-modal fusion transformer for end-to-end autonomous driving.
\newblock In {\em CVPR}, 2021.

\bibitem{rabinovich2007objects}
Andrew Rabinovich, Andrea Vedaldi, Carolina Galleguillos, Eric Wiewiora, and Serge Belongie.
\newblock Objects in context.
\newblock In {\em ICCV}, 2007.

\bibitem{raghu2021vision}
Maithra Raghu, Thomas Unterthiner, Simon Kornblith, Chiyuan Zhang, and Alexey Dosovitskiy.
\newblock Do vision transformers see like convolutional neural networks?
\newblock {\em NeurIPS}, 2021.

\bibitem{ramachandran2019stand}
Prajit Ramachandran, Niki Parmar, Ashish Vaswani, Irwan Bello, Anselm Levskaya, and Jon Shlens.
\newblock Stand-alone self-attention in vision models.
\newblock {\em NeurIPS}, 32, 2019.

\bibitem{recht2019imagenet}
Benjamin Recht, Rebecca Roelofs, Ludwig Schmidt, and Vaishaal Shankar.
\newblock Do imagenet classifiers generalize to imagenet?
\newblock In {\em ICML}, 2019.

\bibitem{ILSVRC15}
Olga Russakovsky, Jia Deng, Hao Su, Jonathan Krause, Sanjeev Satheesh, Sean Ma, Zhiheng Huang, Andrej Karpathy, Aditya Khosla, Michael Bernstein, Alexander~C. Berg, and Li Fei-Fei.
\newblock {ImageNet Large Scale Visual Recognition Challenge}.
\newblock {\em IJCV}, 2015.

\bibitem{shotton2009textonboost}
Jamie Shotton, John Winn, Carsten Rother, and Antonio Criminisi.
\newblock Textonboost for image understanding: Multi-class object recognition and segmentation by jointly modeling texture, layout, and context.
\newblock {\em IJCV}, 81(1):2--23, 2009.

\bibitem{vgg}
Karen Simonyan and Andrew Zisserman.
\newblock Very deep convolutional networks for large-scale image recognition.
\newblock In {\em ICLR}, 2014.

\bibitem{strudel2021segmenter}
Robin Strudel, Ricardo Garcia, Ivan Laptev, and Cordelia Schmid.
\newblock Segmenter: Transformer for semantic segmentation.
\newblock In {\em CVPR}, 2021.

\bibitem{szegedy2016rethinking}
Christian Szegedy, Vincent Vanhoucke, Sergey Ioffe, Jon Shlens, and Zbigniew Wojna.
\newblock Rethinking the inception architecture for computer vision.
\newblock In {\em CVPR}, 2016.

\bibitem{tan2019efficientnet}
Mingxing Tan and Quoc Le.
\newblock Efficientnet: Rethinking model scaling for convolutional neural networks.
\newblock In {\em ICML}, 2019.

\bibitem{mlp_mixer}
Ilya~O Tolstikhin, Neil Houlsby, Alexander Kolesnikov, Lucas Beyer, Xiaohua Zhai, Thomas Unterthiner, Jessica Yung, Andreas Steiner, Daniel Keysers, Jakob Uszkoreit, et~al.
\newblock Mlp-mixer: An all-mlp architecture for vision.
\newblock In {\em NeurIPS}, 2021.

\bibitem{torralba2003contextual}
Antonio Torralba.
\newblock Contextual priming for object detection.
\newblock {\em IJCV}, 53(2):169--191, 2003.

\bibitem{touvron2021resmlp}
Hugo Touvron, Piotr Bojanowski, Mathilde Caron, Matthieu Cord, Alaaeldin El-Nouby, Edouard Grave, Gautier Izacard, Armand Joulin, Gabriel Synnaeve, Jakob Verbeek, et~al.
\newblock Resmlp: Feedforward networks for image classification with data-efficient training.
\newblock {\em arXiv preprint arXiv:2105.03404}, 2021.

\bibitem{deit}
Hugo Touvron, Matthieu Cord, Matthijs Douze, Francisco Massa, Alexandre Sablayrolles, and Herve Jegou.
\newblock Training data-efficient image transformers \& distillation through attention.
\newblock In {\em ICML}, 2021.

\bibitem{tuli2021convolutional}
Shikhar Tuli, Ishita Dasgupta, Erin Grant, and Thomas~L Griffiths.
\newblock Are convolutional neural networks or transformers more like human vision?
\newblock {\em arXiv preprint arXiv:2105.07197}, 2021.

\bibitem{transformer}
Ashish Vaswani, Noam Shazeer, Niki Parmar, Jakob Uszkoreit, Llion Jones, Aidan~N Gomez, \L~ukasz Kaiser, and Illia Polosukhin.
\newblock Attention is all you need.
\newblock In {\em NeurIPS}, 2017.

\bibitem{wang2018non}
Xiaolong Wang, Ross Girshick, Abhinav Gupta, and Kaiming He.
\newblock Non-local neural networks.
\newblock In {\em CVPR}, 2018.

\bibitem{rw2019timm}
Ross Wightman.
\newblock Pytorch image models.
\newblock \url{https://github.com/rwightman/pytorch-image-models}, 2019.

\bibitem{xiao2018unified}
Tete Xiao, Yingcheng Liu, Bolei Zhou, Yuning Jiang, and Jian Sun.
\newblock Unified perceptual parsing for scene understanding.
\newblock In {\em ECCV}, 2018.

\bibitem{YuK15}
Fisher Yu and Vladlen Koltun.
\newblock Multi-scale context aggregation by dilated convolutions.
\newblock In {\em ICLR}, 2016.

\bibitem{yun2019cutmix}
Sangdoo Yun, Dongyoon Han, Seong~Joon Oh, Sanghyuk Chun, Junsuk Choe, and Youngjoon Yoo.
\newblock Cutmix: Regularization strategy to train strong classifiers with localizable features.
\newblock In {\em ICCV}, 2019.

\bibitem{zhang2018mixup}
Hongyi Zhang, Moustapha Cisse, Yann~N. Dauphin, and David Lopez-Paz.
\newblock mixup: Beyond empirical risk minimization.
\newblock In {\em ICLR}, 2018.

\bibitem{zhao2017pyramid}
Hengshuang Zhao, Jianping Shi, Xiaojuan Qi, Xiaogang Wang, and Jiaya Jia.
\newblock Pyramid scene parsing network.
\newblock In {\em CVPR}, 2017.

\bibitem{zhong2020random}
Zhun Zhong, Liang Zheng, Guoliang Kang, Shaozi Li, and Yi Yang.
\newblock Random erasing data augmentation.
\newblock In {\em AAAI}, 2020.

\bibitem{adk20k}
Bolei Zhou, Hang Zhao, Xavier Puig, Sanja Fidler, Adela Barriuso, and Antonio Torralba.
\newblock Scene parsing through ade20k dataset.
\newblock In {\em CVPR}, 2017.

\bibitem{adk20k_2}
Bolei Zhou, Hang Zhao, Xavier Puig, Tete Xiao, Sanja Fidler, Adela Barriuso, and Antonio Torralba.
\newblock Semantic understanding of scenes through the ade20k dataset.
\newblock {\em IJCV}, 127(3):302--321, 2019.

\end{thebibliography}

}

\newpage
\newpage

\setcounter{section}{0}
\renewcommand{\thesection}{\Alph{section}}
\renewcommand{\thefigure}{S\arabic{figure}}
\renewcommand{\thetable}{S\arabic{table}}

\section{Details of Experiments Setup}\label{detail:datasets}
This section provides information on datasets and models used in the main paper with hyper-parameters of the training.

\subsection{Datasets} \label{appendix:datasets}
\textbf{ImageNet-1K.}
ImageNet-1K~\cite{ILSVRC15} is the popular large-scale classification benchmark dataset, and the license is custom for research and non-commercial.
ImageNet-1K consists of 1.28M training and 50K validation images with 1K classes.
We use the training and the validation sets to train and evaluate architectures, respectively.

\textbf{ImageNet-V2.}
ImageNet-V2~\cite{recht2019imagenet} is new test data for the ImageNet benchmark.
Each of the three test sets in ImageNet-V2 comprises 10,000 new images.
After a decade of progress on the original ImageNet dataset, these test sets were collected.
This ensures that the accuracy scores are not influenced by overfitting and that the new test data is independent of existing models.

\textbf{ImageNet-ReaL.}
ImageNet-ReaL~\cite{beyer2020we} develops a more reliable method for gathering annotations for the ImageNet validation set and is under the Apache 2.0 license. 
It re-evaluates the accuracy of previously proposed ImageNet classifiers using these new labels and finds their gains are smaller than those reported on the original labels.
Therefore, this dataset is called the ``Re-assessed Labels (ReaL)'' dataset.

\textbf{ADE20K.}
ADE20K~\cite{adk20k, adk20k_2} is a semantic segmentation dataset, and the license is custom research-only and non-commercial.
This contains over 20K scene-centric images that have been meticulously annotated with pixel-level objects and object parts labels.
There are semantic categories, which encompass things like sky, road, grass, and discrete objects like person, car, and bed.

\textbf{ImageNet-A.}
ImageNet-A~\cite{hendrycks2021nae} is a set of images labeled with ImageNet labels that were created by collecting new data and preserving just the images that ResNet-50~\cite{resnet} models failed to categorize properly. 
This dataset is under the MIT license.
The label space is identical to ImageNet-1K.

\textbf{VQAv2.}
VQA~\cite{VQA} dataset contains open-ended questions about images.
These questions require an understanding of vision, language, and commonsense knowledge to answer.
VQAv2~\cite{balanced_vqa_v2} is the second version of the VQA dataset, which contains 204K COCO images.

\subsection{Models} \label{appendix:models}
Dosovitskiy \etal~\cite{vit} have proposed ViT-B.
Touvron \etal~\cite{deit} have proposed tiny and small ViT architectures named as ViT-Ti and ViT-S.
The ViT architecture is similar to Transformer~\cite{transformer} but has patch embedding to make tokens of images.
Specifically, ViT-Ti/-S/-B have 12 depth layers with 192, 384, and 768 dimensions, respectively.
Heo \etal~\cite{pit} have proposed a variant of ViT by reducing the spatial dimensions and increasing the channel dimensions.
ViTs consist of a patch embedding layer, multi-head self-attention ($\msa$) blocks, multi-layer perceptron ($\mlp$) blocks, and layer normalization ($\norm$) layers.
Our module is the modification of $\mlp$ block.
Our module only requires 1 line modification at the end of the $\mlp$ layer.

\subsection{Hyper-parameters} \label{appendix:hyperparameters}
Touvron \etal~\cite{deit} have proposed data-efficient training settings with strong regularization, such as MixUp \cite{zhang2018mixup}, CutMix \cite{yun2019cutmix}, and random erasing \cite{zhong2020random}.
We adopt the training setting of DeiT \cite{deit} and denote ViT-Ti/-S/-B with $\gs$ module.
We do not use repeated augmentation for ViT-Ti/-S with ours.
For ViT-B with ours, we increase the warmup epochs from 5 to 10 and the drop path.
In distillation, we use the same hyper-parameters except for the drop path of ViT-B with ours to 0.2.

\section{Additional Experiments}
This section provides additional experiments we cannot report due to page limitations.
\begin{figure}
    \centering
    \small
    \includegraphics[width=0.5\linewidth]{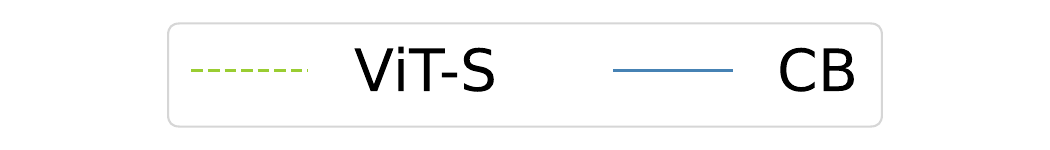}
    
    \begin{subfigure}[b]{0.45\linewidth}
         \centering
         \includegraphics[width=1.0\linewidth]{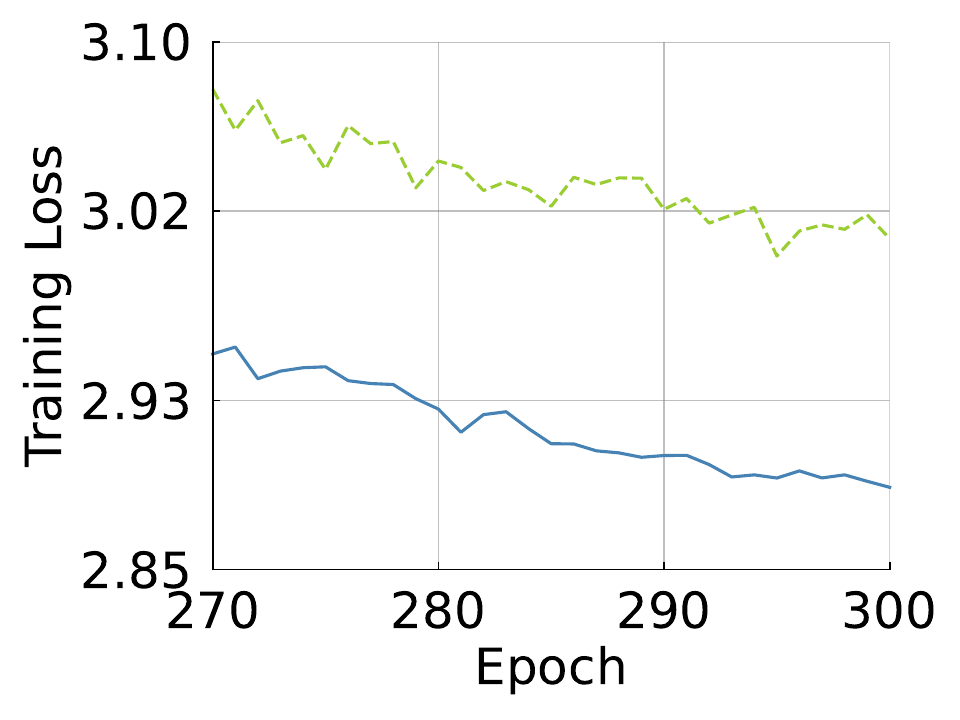}
         \caption{Training Loss}
    \end{subfigure}
    \begin{subfigure}[b]{0.45\linewidth}
         \centering
         \includegraphics[width=1.0\linewidth]{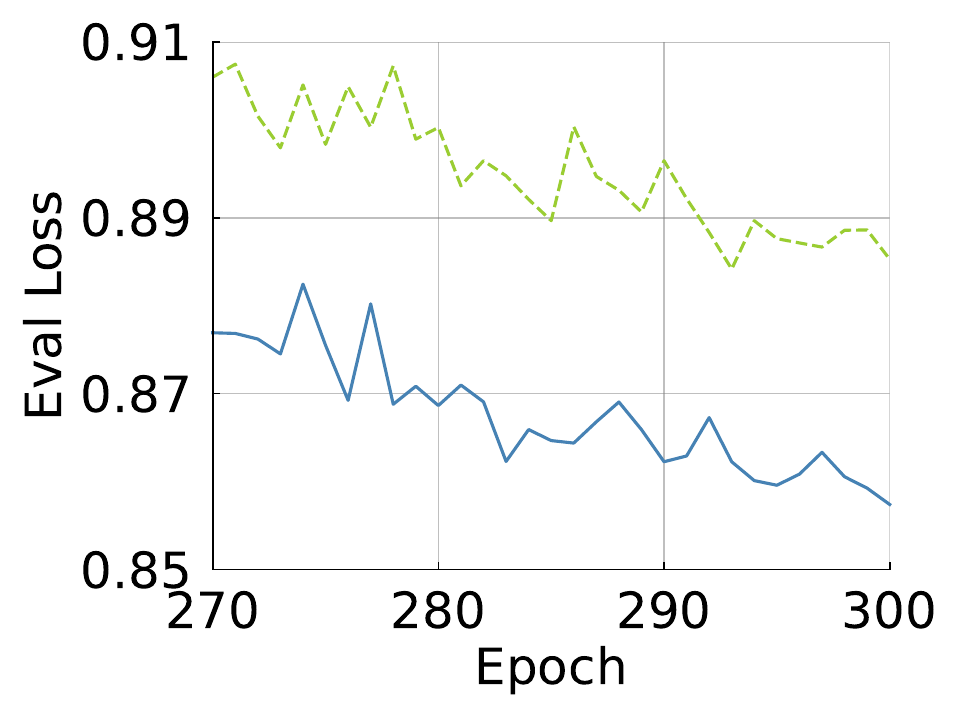}
         \caption{Eval Loss}
    \end{subfigure}
    
    \begin{subfigure}[b]{0.45\linewidth}
         \centering
         \includegraphics[width=1.0\linewidth]{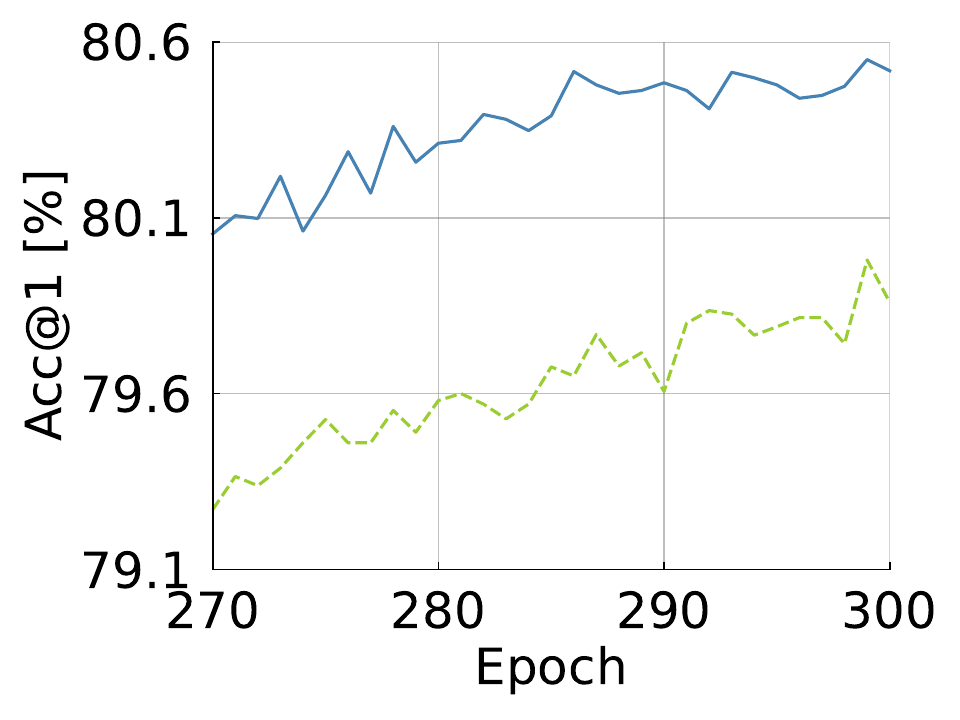}
         \caption{Top-1 Accuracy}
    \end{subfigure}
    \begin{subfigure}[b]{0.45\linewidth}
         \centering
         \includegraphics[width=1.0\linewidth]{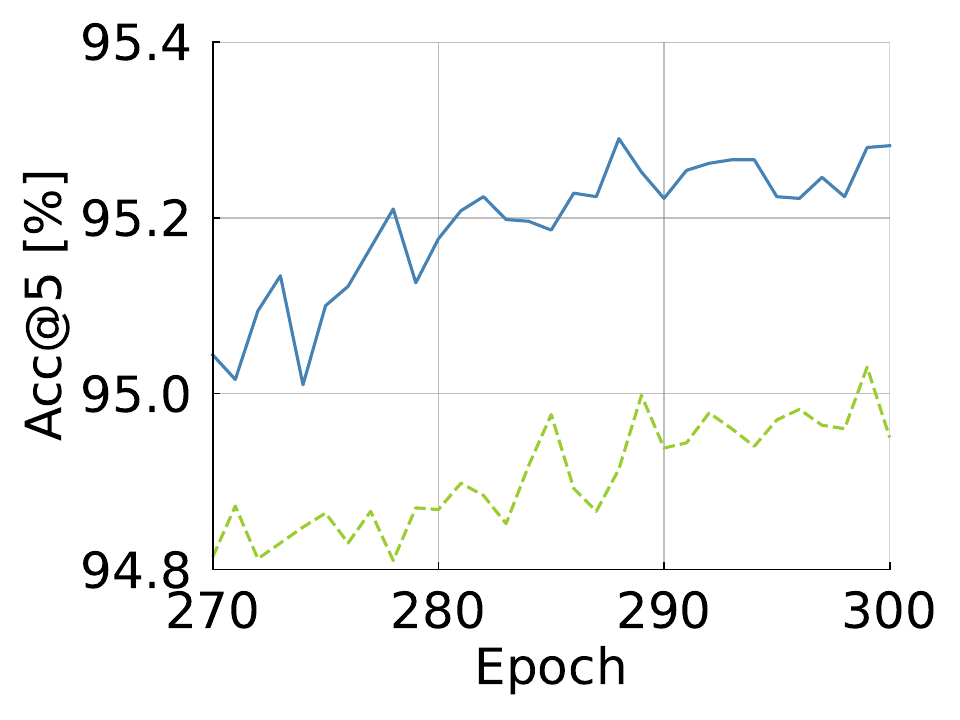}
         \caption{Top-5 Accuracy}
    \end{subfigure}
    
    \caption{\textbf{Training curve of ViT-S with and without $\gs$ module.}
    $\gs$ increases the accuracies across epochs and decreases both training and evaluation losses more. $\gs$ improves the capacity of ViT.}
    \label{fig:training_curve}
\end{figure}
\subsection{Training Curve}
We draw the training curve to see if $\gs$ improves the capacity and convergence of ViTs.
As shown in \Fref{fig:training_curve}, $\gs$ increases the top-1/-5 accuracies across epochs and decreases both training and evaluation losses more.
The curves show that $\gs$ improves the capacity of ViTs.

\begin{table*}
    \centering
    \tabcolsep=0.2cm
    \begin{tabular}{l c c c c c c}
         \toprule
          Architecture & \# Params [M] & FLOPs {[G]} & Acc@1 {[\%]} & Acc@5 {[\%]} & IN-V2 {[\%]} & IN-ReaL [\%] \\ \midrule
         ViT-Ti\distill & 5.9 & 1.3 & 74.5 & 91.9 & 62.4 & 82.1 \\
         $+ \ \gs$ & 5.9 & 1.3 & 74.7 & 92.3 & 62.5 & 82.3 \\
         $+ \ \gss$ & 5.9 & 1.3 & \textbf{75.3} & \textbf{92.5} & \textbf{63.4} & \textbf{82.8} \\
         \midrule
         ViT-S\distill & 22.4 & 4.6 & 81.2 & 95.4 & 69.8 & 86.9 \\
         $+ \ \gs$ & 22.4 & 4.6  & 81.3 & \textbf{95.6} & 70.2 & 87.0 \\
         $+ \ \gss$ & 22.4 & 4.6 & \textbf{81.6} & \textbf{95.6} & \textbf{70.9} & \textbf{87.3} \\
         \midrule
         ViT-B\distill & 87.3 & 17.6 & 83.4 & 96.4 & 72.2 & 88.1 \\
         $+ \ \gs$ & 87.3 & 17.6 & 83.5 & \textbf{96.5} & 72.3 & 88.1 \\
         $+ \ \gss$ & 87.3 & 17.6 & \textbf{83.6} & \textbf{96.5} & \textbf{73.4} & \textbf{88.3} \\
         \bottomrule
    \end{tabular}
    \caption{\small \textbf{ImageNet-1K performance.} We train vision transformer architectures~\cite{vit, deit} with $\gs$ and $\gss$ and evaluate the accuracy on ImageNet-1K~\cite{deng2009imagenet}, ImageNet-V2~\cite{recht2019imagenet}, and ImageNet-ReaL~\cite{beyer2020we}. 
    \textbf{Bold} is the best number at each row. 
    Our module improves all the metrics incurring negligible extra computational costs.
    }
    \label{table:imnet}    
    \vspace{-1em}
\end{table*}

\begin{table}[]
    \centering
    \small
    \tabcolsep=0.2cm
    \begin{tabular}{l c c c}
         \toprule
          Architecture & Occ [\%] & ImageNet-A [\%] & FGSM [\%]\\
         \midrule
         ViT-S\distill & 74.6 & 21.5 & 11.8 \\
         $+ \ \gss$ & 75.1 & 22.6 & 13.0 \\
         $+ \ \gs$ & \textbf{75.1} & \textbf{23.6} & \textbf{15.5} \\
         \bottomrule
    \end{tabular}
    \caption{\textbf{Robustness evaluation.} We evaluate ViT-S\distill with $\gs$ and $\gss$ on center occlusion (Occ), ImageNet-A, and fast sign gradient method (FGSM) attack. Ours shows improved robustness across the board against ViT-S\distill.}
    \label{table:robo}
\end{table}

\subsection{Distilled Performance}
We follow the specification of ViT-Ti/-S from DeiT~\cite{deit}.
As shown in \Tref{table:imnet}, our modules $\gs$ and $\gss$ improve performance compared to the distilled ViTs consistently.
Table~\ref{table:robo} shows the results of the robustness benchmark in ViT-S\distill.
$\gss$ increases 0.5, 1.1, and 2.2 of Occ, ImageNet-A, and FGSM, respectively.
$\gs$ does 0.5, 2.1, and 4.7, respectively.

\subsection{Architecture generalizability.}
To show further applicability of our module, 
We compare PVT, LocalViT-Ti, PVTv2, and Swin trained w/ and w/o ours on ImageNet-1K;
we set the default epochs to 120.\footnote{We set 150 epochs for LocalViT.}
PVT uses a pyramid structure as CNN backbones,
LocalViT-Ti puts the convolution layer (local operation),
PVTv2 is the improved version of the PVT placing convolutional layer,
and Swin employs a hierarchical structure and local attention.
\Tref{tab:other-architectures} shows that our module consistently improves performance. 

\begin{table}[]
    \centering
    \small
    \begin{tabular}{c| c c | c}
        \toprule
        Model & Hierarchy & Local & ACC@1[\%] \\
        \midrule
        PVT-S & \checkmark & & 76.7\\
        + Ours & \checkmark & & \textbf{77.3} \\\midrule
        LocalViT-Ti &  & \checkmark & 69.4\\
        + Ours &  & \checkmark & \textbf{72.5} \\\midrule
        PVTv2-B1 & \checkmark & \checkmark & 76.4\\
        + Ours & \checkmark & \checkmark & \textbf{76.5} \\\midrule
        Swin-Ti & \checkmark & \checkmark & 79.3\\
        + Ours & \checkmark & \checkmark & \textbf{79.5} \\\bottomrule
    \end{tabular}
    \vspace{-.5em}
    \caption{\textbf{ImageNet-1K results with hierarchical ViTs.} 
    We further report the results of training on ImageNet-1K.}
    \label{tab:other-architectures}
\end{table}

\begin{figure}
        \centering
        
        \begin{subfigure}[b]{0.99\linewidth}
            \centering
            \includegraphics[width=1.0\linewidth]{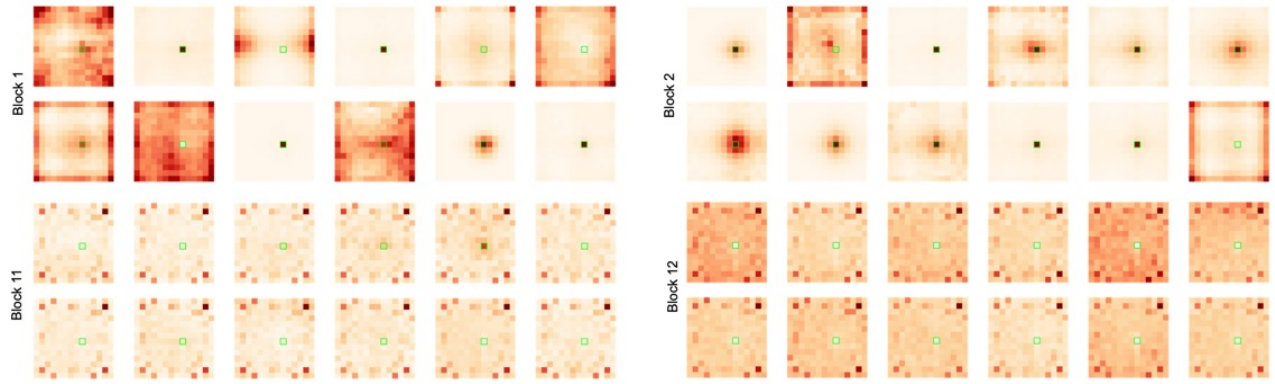}
            \caption{He \etal~\cite{he2021pruning}}
            \label{fig:he}
        \end{subfigure}
        
        \begin{subfigure}[b]{0.99\linewidth}
            \centering
            \includegraphics[width=1.0\linewidth]{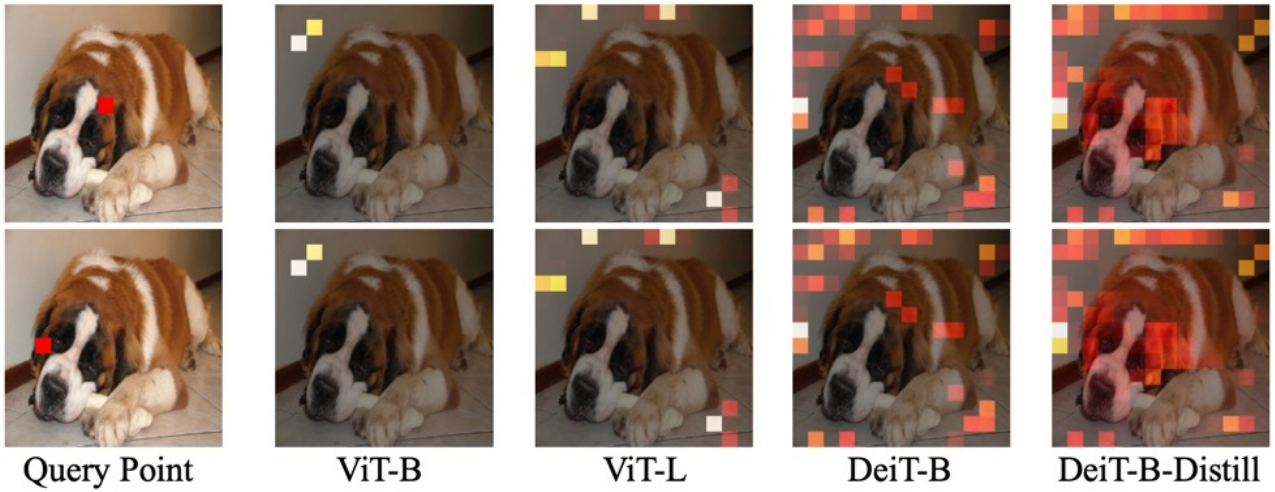}
            \caption{Ma \etal~\cite{ma2022close}}
            \label{fig:ma}
        \end{subfigure} 
        \vspace{-2mm}
        \caption{\textbf{Visualization from He \etal~\cite{he2021pruning} and Ma \etal~\cite{ma2022close}.}
        (a) He \etal~\cite{he2021pruning} visualized the attention of DeiT-B 12 heads in shallow and deeper layers.
        (b) Ma \etal~\cite{ma2022close} visualized  the attention of ViTs given the query point.}
        \label{fig:visualization}
\end{figure}

\subsection{Discussion on Attention Visualization}
The visualization of the attention map is not the first attempt.
As shown in \Fref{fig:visualization}, He \etal~\cite{he2021pruning} and Ma \etal~\cite{ma2022close} visualized the attention across layers or architectures.
He \etal~\cite{he2021pruning} reported that the deeper layers attend the dense global regions, and shallow layers attend the sparse local regions in DeiT-B.
Ma \etal~\cite{ma2022close} reported that the attention weights are sparse in DeiT-B compared to DeiT-B-Distill.
Despite the fact that they analyze the same architecture, DeiT-B, He \etal and Ma \etal argued the different statements on sparsity.
They use different criteria depending on what they want to compare in \emph{relative ways}.
Thereby, it has been open discussion in the community about sparsity characteristics of attention maps in Transformers due to those subjective visualization-based analyses.

Based on the visualization, the statements of He \etal and Ma \etal are conditional on the reference of an attention visualization.
These conditional statements can be changed by choosing a different reference; thus, our observation is not contrary to theirs.
To compare more general ways, we employ an objective measure, \ie, entropy measure; high entropy implies dense interactions and vice versa.
It is our contribution.
Our entropy analysis in Sec. 3 supports the visualization (Fig. 10), where $\gs$ lowers the entropy and helps $\msa$ attend to more informative signals.

\begin{table}
    \small
    \centering
    \resizebox{0.5\linewidth}{!}{
    \begin{tabular}{c c}
         \toprule
         Architecture & Accuracy\\
         \midrule
         ViT-S & 74.73\\
         + $\gs$ & \textbf{75.43}\\
         \bottomrule
    \end{tabular}}
    \caption{\textbf{Accuracy on CIFAR-100.} We train ViT-S from scratch on CIFAR-100 dataset. $\gs$ increases the accuracy by 0.7\%p.}
    \label{table:cifar100}
\end{table}
\subsection{Classification on CIFAR-100}
We train ViT-S from scratch on CIFAR-100.
CIFAR-100 consists of 50,000 training and 10,000 validation images with 100 classes.
\Tref{table:cifar100} shows the accuracy on CIFAR-100.
$\gs$ improves the performance by 0.7\%p more.

\subsection{Discussion on Position of $\gs$}
We conclude the position of $\gs$ to the end of the $\mlp$ block.
We provide our intuition and discussion about Table 3-(b), which shows performance depends on $\gs$ position in the $\mlp$ block.

\paragraph{Gradient signals.}
We think that the gradient signals are dependent on the position of $\gs$.
For simplicity, we assume a single layer composed of the $\msa$ and $\mlp$ block.
Let the $\mlp$ layer consist as follows: $\mathtt{<Front> - FC layer - <Mid> - FC layer - <End>}$.
\begin{itemize}
    \item Case 1, $\Front$: If $\gs$ is located at $\Front$, the subsequent weights in the corresponding $\mlp$ block cannot receive the gradient signals during training.
    \item Case 2, $\End$: If $\gs$ is located at $\End$, the preceding weights in the $\mlp$ block are updated by the gradient signals by uniform attention.
\end{itemize}

\paragraph{Why is the improvement of $\Mid$ and $\End$ similar?}
There is no non-linear function (e.g., GELU) between $\Mid$ and $\End$ positions. 
Since uniform attention is the addition of a globally averaged token, the output is identical wherever $\gs$ is located at $\Mid$ and $\End$. 
Therefore, the accuracy of both positions is similar.
Nonetheless, $\gs$ at $\End$ achieves a bit higher top-5 accuracy than $\gs$ at $\Mid$.
As aforementioned, we suspect the $\End$ position provides the gradient induced by uniform attention to weights of the MLP block.

\begin{table}
    \small
    \centering
    \resizebox{1.0\linewidth}{!}{
    \begin{tabular}{l c c c c c}
         \toprule
         \multirow{2}[2]{*}{Module} & \multicolumn{2}{c}{Position} & \multirow{2}[2]{*}{FLOPs {[M]}} & \multirow{2}[2]{*}{Acc@1 {[\%]}} & \multirow{2}[2]{*}{Acc@5 {[\%]}} \\\cmidrule(lr){2-3}
          & $\mlp$ & $\msa$ & & \\
         \midrule
         ViT-S & \xmark & \xmark & 1260 & 79.9 & 95.0 \\ 
         \midrule
          \multirow{3}{*}{+$\gs$} & \cmark & \xmark & +0.9 & \textbf{80.5} & \textbf{95.3} \\
          & \xmark & \cmark & +0.9 & 80.1 & 95.0\\
          & \cmark & \cmark & +1.8 & 80.1 & 95.0\\
         \midrule
         \multirow{3}{*}{+$\gsmul$} & \cmark & \xmark & +0.9 & \textbf{80.4}  & \textbf{95.1} \\
          & \xmark & \cmark & +0.9 & 80.0 & 95.0\\
          & \cmark & \cmark & +1.8 & 80.0 & 95.0\\
         \midrule
         \multirow{3}{*}{+$\gshyb$} & \cmark & \xmark & +1.8 & \textbf{80.5} & 95.0 \\
          & \xmark & \cmark & +1.8 & 80.4 & \textbf{95.3} \\
          & \cmark & \cmark & +3.6 & - & -\\
         \bottomrule
    \end{tabular}}
    \caption{\textbf{Performance of ViT-S with $\gs, \gsmul$, and $\gshyb$.} 
    We train ViT-S with $\gs, \gsmul$, and $\gshyb$ on ImageNet-1K training set and evaluate top-1/-5 accuracies on the validation set. 
    We vary the position of our modules at $\mlp$, $\msa$, and both. 
    \textbf{Bold} is the best number at each row.
    }
    \label{table:pos_blocks}
\end{table}

\begin{table}
    \centering
    \small
    \resizebox{1.0\linewidth}{!}{
    \begin{tabular}{l c c c c l l}
         \toprule
         \multirow{2}[2]{*}{Module} & \multicolumn{3}{c}{Position} & \multirow{2}[2]{*}{FLOPS {[M]}} & \multirow{2}[2]{*}{Acc@1 {[\%]}} & \multirow{2}[2]{*}{Acc@5 {[\%]}} \\ \cmidrule(lr){2-4}
          & $\Front$ & $\Mid$ & $\End$ & &  \\ \midrule
         ViT-S & \xmark & \xmark & \xmark & 1260 & 79.9 & 95.0 \\ \midrule
          \multirow{3}{*}{+$\gs$} & \cmark & \xmark & \xmark & +0.9 & 79.9 & 94.8 \\
          & \xmark & \cmark & \xmark & +3.6 &  \textbf{80.5} & 95.2 \\
          & \xmark & \xmark & \cmark & +0.9 & \textbf{80.5} & \textbf{95.3} \\ 
          \midrule
          \multirow{3}{*}{+$\gsmul$} & \cmark & \xmark & \xmark & +0.9 & 80.3 & 94.9 \\
          & \xmark & \cmark & \xmark & +3.6 & 80.2 & \textbf{95.1} \\
          & \xmark & \xmark & \cmark & +0.9 & \textbf{80.4} & \textbf{95.1} \\ \midrule
          \multirow{3}{*}{+$\gshyb$} & \cmark & \xmark & \xmark & +1.8 & \textbf{80.5} & 95.0\\
          & \xmark & \cmark & \xmark & +7.3 & 80.1 & \textbf{95.1} \\
          & \xmark & \xmark & \cmark & +1.8 & 80.3 & 95.0\\
          \bottomrule
    \end{tabular}}
    \caption{\textbf{Performance of ViT-S with different positions in $\mlp$ block.} $\mlp$ has following schematic: $\mathtt{<Front> - FC layer - <Mid> - FC layer - <End>}$. 
    We insert $\gs, \gsmul$, and $\gshyb$ at $\Front$, $\Mid$, and $\End$ and evaluation on ImageNet-1K. 
    \textbf{Bold} is the best number at each row.
    }
    \label{table:pos_mlp}
\end{table}

\subsection{Utilizing the Class Token}
Since the class token evolves by interacting with entire tokens for tasks, we think that the class token could be utilized to complement spatial interactions of attention.
We propose two additional baselines employing the class token.

The first one is the multiplication of the class token with each visual token, similar to the gating mechanism.
We denote the first as $\gsmul$ and formalize it as follows:
$\gsmul (\bx_i) = \bx_i (\bx_{0} + \mathbf{1}) \text{ for every token $i$}$, where $\bx_0$ is the class token and $\mathbf{1}$ is one vector.
The second one is the combination of the class and average token denoted as $\gshyb$:
$\gshyb (\bx_i) = \bx_i \bx_{0} +  \gs(\bx_i) \text{ for every token $i$}$.
These modules are also parameter-free and computation efficient.

Firstly, we analyze the positions of $\mlp$ and $\msa$; we locate the modules at the end of blocks.
Table \ref{table:pos_blocks} lists FLOPs and validation accuracy of $\mlp$ and $\msa$.
Both $\gsmul$ and $\gshyb$ improve the top-1 accuracy regardless of positions except for the failure case of $\gshyb$ at both $\mlp$ and $\msa$ layers.
These modules have the best top-1 accuracy at $\mlp$, consistent with our $\gs$ module.

We investigate different positions in an $\mlp$ layer with $\gsmul$ and $\gshyb$.
Table \ref{table:pos_mlp} lists FLOPs and validation accuracies of $\Front$, $\Mid$, and $\End$.
The best accuracy occurs at $\End$ for our $\gs$ module and $\gsmul$ and $\Front$ for $\gshyb$.
At the best positions of respective modules, our $\gs$ module achieves 0.1\%p higher top-1 accuracy than $\gsmul$ and demands half of the FLOPs than $\gshyb$.

\end{document}